\documentclass[letterpaper]{article} 
\usepackage[submission]{aaai25}  
\usepackage{times}  
\usepackage{helvet}  
\usepackage{courier}  
\usepackage[hyphens]{url}  
\usepackage{graphicx} 
\usepackage{natbib}  
\usepackage{caption} 
\usepackage{algorithm}
\usepackage{algorithmic}

\usepackage{newfloat}
\usepackage{listings}
\DeclareCaptionStyle{ruled}{labelfont=normalfont,labelsep=colon,strut=off} 
\floatstyle{ruled}
\newfloat{listing}{tb}{lst}{}
\floatname{listing}{Listing}
\usepackage{xcolor}         
\usepackage{amsthm}
\usepackage{amssymb}
\usepackage{enumitem}       
\usepackage{eqparbox}       
\usepackage{paralist}       
\usepackage{multirow} 
\usepackage{booktabs}       
\usepackage[font={scriptsize}]{subcaption}
\usepackage{makecell} 
\usepackage{array} 
\usepackage{hyperref}

\newcommand{\longcomment}[1]{\\ \hfill \textcolor{gray}{\(\triangleright\)} \parbox[t]{.8\linewidth}{\textcolor{gray}{#1}}}

\theoremstyle{definition}
\newtheorem{definition}{Definition}[section]
\newcommand{\app}{{Suppl~}}

\newcommand{\MethodName}{\emph{TEAL}}
\newcommand{\floor}[1]{\lfloor #1 \rfloor}

\newcommand{\myparagraph}[1]{\smallskip\noindent\textbf{#1}\hspace{.3cm}}

\definecolor{orange}{rgb}{1, 0.75, 0.35}
\definecolor{blue}{rgb}{0.36, 0.54, 1}
\definecolor{pink}{rgb}{1.0, 0.4, 0.4}

\newcommand{\T}{\mathcal{T}}

\newcommand{\sR}{{\mathbb{R}}}

\newcommand{\X}{{\mathcal{X}}}
\newcommand{\mem}{{\mathcal{M}}}

\usepackage{amsmath}
\DeclareMathOperator*{\argmax}{argmax}

\title{TEAL: New Selection Strategy for Small Buffers  in Experience Replay \\ Class Incremental Learning}
\author{
    Shahar Shaul-Ariel\textsuperscript{\ensuremath\dagger}, Daphna Weinshall\textsuperscript{\ensuremath\dagger}\\
    School of Computer Science \& Engineering\textsuperscript{\ensuremath\dagger} \\
    The Hebrew University of Jerusalem \\  Jerusalem 91904, Israel\\
    {\tt\small \{shahar.ariel1,daphna\}@mail.huji.ac.il}\\
}

\begin{document}

\maketitle

\begin{abstract}
Continual Learning is an unresolved challenge, whose relevance  increases when considering modern applications. Unlike the human brain, trained deep neural networks suffer from a phenomenon called \emph{catastrophic forgetting}, wherein they progressively lose previously acquired knowledge upon learning new tasks. To mitigate this problem, numerous methods have been developed, many relying on the replay of past exemplars during new task training.  However, as the memory allocated for replay decreases, the effectiveness of these approaches diminishes. On the other hand, maintaining a large memory for the purpose of replay is inefficient and often impractical. Here we introduce \MethodName, a novel approach to populate the memory with exemplars, that can be integrated with various experience-replay methods and significantly enhance their performance with small memory buffers. We show that \MethodName\ enhances the average accuracy of existing class-incremental methods and outperforms other selection strategies, achieving state-of-the-art performance even with small memory buffers of 1-3 exemplars per class in the final task.
This confirms our initial hypothesis that when memory is scarce, it is best to prioritize the most typical data. Code is available at this \href{https://github.com/shahariel/TEAL}{https URL}.
\end{abstract}

\section{Introduction}
\label{sec:intro}

With the recent advances in deep neural networks, there has been a growing research interest in incremental learning. The need to integrate new task knowledge into an already trained network has become increasingly important, especially considering the time-consuming nature of training on large datasets. Retraining the network from scratch on both the original and new task data is often impractical, and access to the original training data may be limited or unavailable. In this context, \emph{catastrophic forgetting}, as described by \citet{mccloskey1989catastrophic}, can be particularly severe.

\begin{figure}[H]
    \center
    \includegraphics[width=\linewidth]{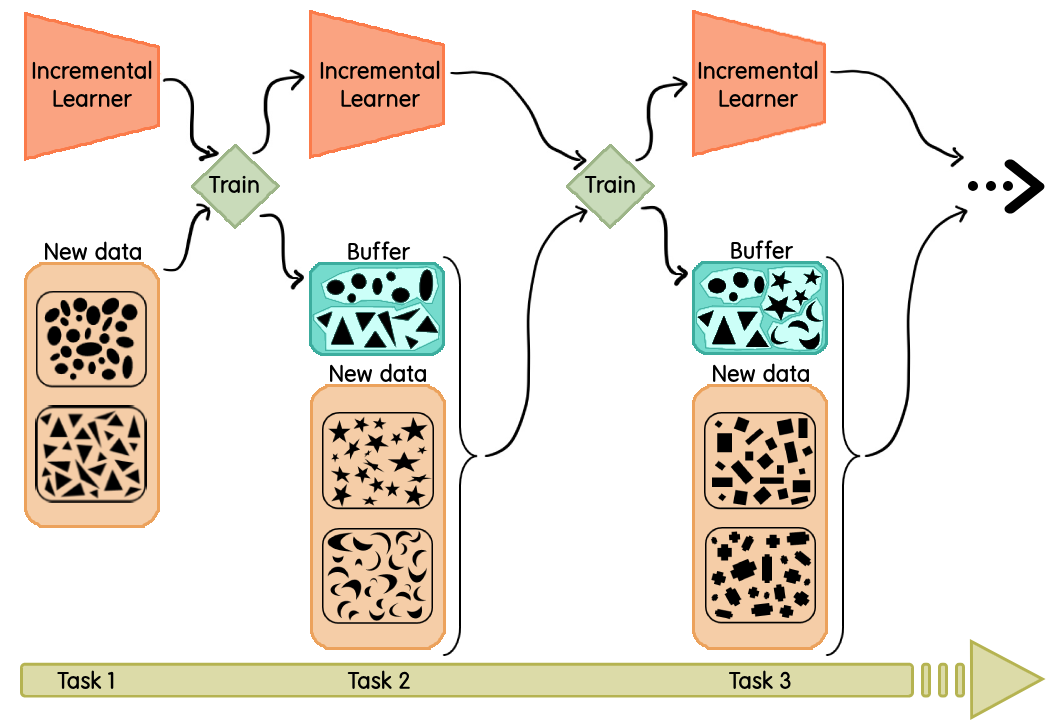}
    \caption{Illustration of CIL with Experience Replay.
\label{fig:cil_illustration}}
\end{figure}

To address this challenge, various methods have been developed across different frameworks. \citet{van2022three} categorizes these methods into three types: task-incremental, domain-incremental, and class-incremental learning. The fundamental idea behind incremental learning is that a model must sequentially learn tasks, one after the other. Among these approaches, Class-Incremental Learning (CIL) is recognized as the most challenging. Here, each task introduces new classes, and the model must accurately identify the class of each input without access to the corresponding task ID, see Fig.~\ref{fig:cil_illustration}.

Incremental learning can be approached in various ways, each with its own assumptions and configurations. In this paper, we follow the constrained setup outlined by \citet{de2021continual}, which does not depend on task boundaries during either training or testing. This approach maintains a fixed memory size throughout incremental training, ensuring it stays within a predefined limit set from the beginning. At each incremental step, new exemplars can be added to the memory buffer only after sufficient space has been vacated.

We focus our attention on a prevalent and rather successful framework called Experience Replay (ER), which involves storing a set of exemplars in memory and reusing them for rehearsal purposes while training on new tasks. Within this framework, several strategies exist for selecting which exemplars to retain in memory. 
Not surprisingly, the smaller the memory buffer is, the less effective the strategy is at mitigating catastrophic forgetting. This leads us to the following question: Are these strategies necessarily optimal for all memory sizes? In other words, can different strategies be found suitable for different sizes of the memory buffer?

In active learning, it has been shown (empirically and formally) that when the number of labeled examples is small, it is best to choose the most typical examples for training \citep{hacohen2022active}. In the context of CIL and when the memory buffer is too small to truly represent the distribution of each class, we propose to adopt this strategy for the selection of points that are intended to populate the replay memory buffer. In other words, when the buffer is considerably small, it should contain representative exemplars and thus retain a more significant fraction of the previously acquired knowledge.

Our proposed method \MethodName, Typicality Election Approach to continual Learning, is primarily targeted at scenarios with small buffers. While the use of small buffers may seem too strict, there are situations where maintaining a large memory for replay is simply not feasible. For example, applications running on mobile devices face severe memory constraints and could greatly benefit from methods specifically designed for small buffers. 

Accordingly, \MethodName\  aims to identify a set of representative exemplars that also exhibit diversity. An exemplar is deemed representative if its likelihood, when considered within the distribution of all points, is high. To ensure diversity, data clustering is leveraged. 
Central to the success of our approach is the ability to derive an appropriate latent space, where data clustering and likelihood estimation can be reliably obtained.
Ideally, any mechanism for memory population can be combined with any competitive Experience-Replay-Incremental-Learning (ER-IL) method in which this mechanism is a separate module. In accordance, we evaluate our approach in Section~\ref{sec:results} by considering alternative ER-IL methods, where we replace their native mechanism for buffer population by \MethodName.
The method, methodology and its experimental evaluation are described in the rest of this paper, with emphasis on the enhancements \MethodName\ offers to various existing class-incremental methods.

\myparagraph{Related work}
Several approaches exist for incremental learning \citep[see][]{de2021continual}, including Experience Replay (ER), Generative Replay (GR) \citep{shin2017continual}, Parameter Isolation, and Regularization. Similar to ER, GR replays data from previous tasks during new task training, but uses a generative model to create new samples instead of retaining exemplars seen by the model \citep{choi2021dual, gao2023ddgr, gautam2024generative}. Parameter Isolation assigns distinct parameters to each task to reduce forgetting \citep{mallya2018packnet} by fixing parameters assigned to previous tasks, while Regularization-based methods \citep{li2017learning} incorporate additional terms into the loss function to retain prior knowledge while learning from new data. 

These methods differ in the ways they utilize memory: ER stores exemplars, GR stores a generative model, and Parameter Isolation stores task-specific parameters. Regularization-based methods do not rely on memory. This diversity makes it challenging to compare methods directly. In particular, we note that since generative models tend to be very large, GR is hardly suitable to the domain of small memory buffers addressed here. A similar concern can be raised concerning Parameter Isolation methods. Hence, our focus in this paper is on ER.

Note that while few-shot incremental learning \citep{tian2024survey} may appear similar to our work on CIL with a small memory buffer, there is a significant difference. Specifically, we assume that the data for new tasks is initially sufficient for effective learning, whereas few-shot incremental learning inherently deals with a scarcity of labeled samples.

When comparing strategies for the population of the memory buffer, \citet{masana2022class} demonstrate that the most successful strategies are either random-sampling or \textit{Herding} as defined by \citet{welling2009herding}. The latter strategy involves retaining a set of exemplars whose mean is closest to the class mean. Another successful strategy presented by \citet{bang2021rainbow} leverages classification uncertainty and data augmentation to enhance the diversity of data instances (\textit{Uncertainty}). Other selection strategies, such as GSS \citep{aljundi2019gradient} and selecting the exemplars with the highest entropy of the softmax outputs \citep{chaudhry2018riemannian}, have been shown in previous studies to be inferior to random sampling and \textit{Herding} \cite{masana2022class, prabhu2023computationally}. Therefore, we do not include them in our comparative empirical evaluation.

Recently, \citet{hacohen2024forgetting} proposed a selection strategy called Goldilocks, which retains exemplars learned at an intermediate pace, accommodating various buffer sizes. Since the results presented in this paper are primarily within a task-incremental framework (which yields higher average accuracy) and the paper does not provide code, we do not include this method in our comparisons.

\myparagraph{Summary of contribution}
We present \MethodName, a novel strategy designed to select effective exemplars for small memory buffers. \MethodName\ is seamlessly integrated with a number of SOTA replay-based class-incremental learning method, and significantly enhances their performance.

\section{Problem formulation}
\label{sec:theory}

A CIL problem $\T$ consists of a sequence of $T$ tasks. Each task $t\in T$ contains a set of classes $C^t=(c^1,\dots,c^{n_t})$ and labeled samples from these classes $X_t=(X^1,\dots,X^{n_t})$, where $X^i=\{(x_1,i),\dots,(x_{m^i},i)\}$. The tasks do not share classes, i.e., $C^{t_i}\cap C^{t_j}=\emptyset~\forall i\ne j$.
We denote by $N^t$ the total number of classes in all tasks up to task $t$: $N^t=\sum_{i=1}^t{n_i}$.

We consider the scenario where there is access to a memory buffer  $\mem$, which has a fixed size throughout the learning of $\T$. 
An incremental learner is a learning model (we consider only deep neural networks) that is trained sequentially on the tasks. Accordingly, the training on task $t$ is performed on data $X_t \cup \mem$, where $\mem$ contains stored exemplars from tasks $1,\dots,t-1$.
At test time, a CIL method is required to classify a given example into its predicted class while considering all previously seen classes. 

Our proposed method \MethodName\ is described in Section~\ref{sec:method}. It aims to address only one component of the general Incremental Learning (IL) problem, namely, how to populate the memory buffer at the end of each IL iteration. Accordingly, after task $t$, \MethodName\ should select for each seen class $n=\frac{|\mem|}{N_t}$ exemplars to populate memory buffer $\mem$.

\section{Our method: \MethodName}
\label{sec:method}

\begin{figure}[thb]
    \center
    \includegraphics[width=\linewidth]{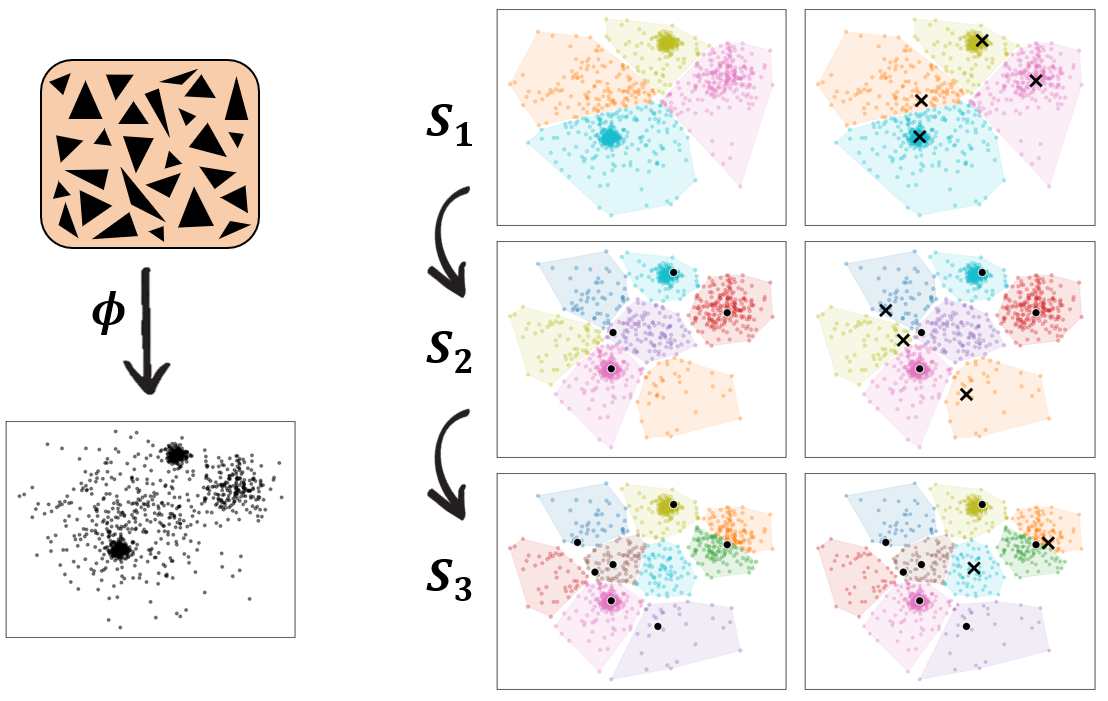}
    \caption{Illustration of \MethodName's iterative class selection process, which establishes a priority order for the selected set. Initially, an embedding space is generated separately for each class (shown on the left). Samples are then selected iteratively with $s_1=4$, $s_2=7$, and $s_3=9$ (see text for details). Each row on the right panel represents one iteration: the left image displays the $s_i$ clusters (obtained using K-means) with the previously selected points $S_{i-1}$ marked with 'o', while the right image shows the updated set $S_i$, with newly selected samples marked with 'X'. In the last iteration 3 clusters remain uncovered, but only 2 samples are selected, leaving the red cluster uncovered by $S_3$.}
\label{fig:teal_example}
\end{figure}

We begin with a definition of point \emph{typicality}.
\begin{definition}[Typicality] 
For each exemplar $x$, let
\[Typicality(x) = \left(\frac{1}{K}\sum_{x_i \in K\textrm{-}NN(x)}{\|x-x_i\|_2}\right)^{-1}\]
where $K$ is a fixed number of nearest neighbors\footnote{We use $K=20$, but other options yield similar results.}. 
\end{definition}
\noindent
The essence of \MethodName\  is to populate the memory buffer with points that are both \emph{typical} (as captured by the definition above) and \emph{diverse}.

After training on task $t$, the incremental learner has to update the memory buffer $\mem$ with data from the new classes $\{X^{N^{t-1}+1},\dots,X^{N^t}\}$. 
Given the fixed size of $\mem$, the learner must first reduce the number of exemplars already in the buffer to accommodate new ones from the new classes. For each class, this involves repeatedly removing examples from the buffer after each IL step.

To address this challenge, \MethodName\ maintains a priority list of selected exemplars from each class, which reflects the order in which they should be removed from $\mem$. This approach allows the learner to remove the least typical points from the buffer when new classes emerge. The list is structured so that the most typical and diverse exemplars, which should be retained as long as possible, are positioned at the top, while those slated for earlier removal are positioned at the bottom.

More specifically, consider a fixed class, and let $n$ denote the number of exemplars assigned to that class. \MethodName\ repeatedly selects a small fraction of $n$, generating a sequence of subsets $S_1 \subseteq S_2 \subseteq \dots \subseteq S_k$, where $s_i = |S_i|$ and $s_1 \leq \dots \leq s_k = n$. The sizes ${s_i}$ define the selection pace, indicating the rate at which exemplars are accumulated. The inclusion relation induces the priority order: points in $S_k \setminus S_{k-1}$ are removed first, while points in $S_1$ are removed last. This process is illustrated in Fig.~\ref{fig:teal_example}.

To initialize the selection process, we need a suitable embedding space for the class exemplars. To this end we train a deep model on all the available training data, and use activations in its penultimate layer as a representation for the new classes. Subsequently, the selection process in the $i^{th}$ iteration involves 2 steps (see pseudocode in Alg.~\ref{alg:construct_exemplar_set}):
\begin{description}[leftmargin=.8cm]
\item[Step 1: Clustering.] In order to ensure diversity, we seek typical exemplars from different regions of the embedding space. When constructing set $S_i$, we achieve this by dividing the set of labeled points into $s_{i}$ clusters using K-Means \cite{lloyd1982least}.
\item[Step 2: Typicality.] We then select the most typical point from each of the $s_i-s_{i-1}$ largest uncovered clusters, where an uncovered cluster is a cluster from which no point has already been selected. 
\end{description}

\begin{algorithm}[h]
\caption{\textit{ConstructExemplarSet}}
\label{alg:construct_exemplar_set}
    \footnotesize
    \textbf {Input:} a set of exemplars from class $c$ $X^c$, number of examples to choose $n$ 
    \\
    \textbf {Require:} current feature function $\phi:\X\to\sR^d$, iterations pace $s_1,\dots,s_k=n$ 
    \\
    \textbf {Output:} a set of \textit{n} exemplars $\mem_c \subseteq X^c$ 
    \begin{algorithmic}[1]
    \STATE $X^c_{emb} \leftarrow \phi(X^c)$
    \STATE $\mem_c \leftarrow \emptyset$
    \FORALL{$i = 1,\dots,k$}
        \STATE $C_1,\dots,C_{s_i} \leftarrow$ clustering\_algorithm$(X^c_{emb},\ s_i)$ \longcomment{$|C_1|\geq\dots\geq|C_{s_i}|$}
        \FORALL{$j = 1,\dots,s_i-s_{i-1}$}
            \IF{$C_j$ is uncovered }
                \STATE add $\argmax_{x\in C_j} \{Typicality(x)\}$ to $\mem_c$
            \ENDIF
        \ENDFOR
    \ENDFOR
    \STATE {\bfseries return} $\mem_c$ 
\end{algorithmic}
\end{algorithm}

The integration of \MethodName\ into a general class-incremental algorithm is described in Alg.~\ref{alg:incremental_train}.

\begin{algorithm}[h]
\caption{\textit{IncrementalTraining}}
\label{alg:incremental_train}
    \footnotesize
    \textbf {Input:} training examples $X_t$ 
    \\
    \textbf {Require:} current exemplar sets $\mem = (\mem_1,\dots,\mem_{N^{t-1}})$, current model parameters $\theta^{t-1}$ 
    
    \begin{algorithmic}[1]
    \STATE $\theta^t \leftarrow \textrm{training\_model}(X_t,\mem;\theta^{t-1})$
    \STATE $n \leftarrow \frac{|\mem|}{N^t}$ 
    \FOR{$c \in \{1,\dots,N^{t-1}\}$}
        \STATE $\mem_c \leftarrow$ first $n$ exemplars in $\mem_c$ 
    \ENDFOR
    \FOR{$c \in \{N^{t-1}+1,\dots,N^t\}$}
        \STATE $\mem_c \leftarrow ConstructExemplarSet(X^c,n,\theta^t)$
    \ENDFOR
    \STATE $\mem \leftarrow (\mem_1,\dots,\mem_{N^t})$
\end{algorithmic}
\end{algorithm}

\section{Empirical evaluation}
\label{sec:results}

We report two settings: \begin{inparaenum}[(i)] \item \emph{Integrated:} we evaluate the beneficial contribution of  \MethodName\ to competitive ER-IL methods, by replacing their native selection strategy (\textit{vanilla} version) with \MethodName\ (Section~\ref{sec:integrated-TEAL}); \item \emph{Alternative selection strategies:} we evaluate the beneficial contribution of  \MethodName\ as compared to other selection strategies
(Section~\ref{sec:herding-TEAL}). \end{inparaenum} In both cases, we incrementally train a deep model with a fixed-size replay buffer and monitor the average accuracy defined below upon completion of each task $t$. 

\subsection{Methodology}
\label{sec:results_methodology}

The majority of the experiments are conducted using the open-source Continual Learning library Avalanche \citep{lomonaco2021avalanche}, with the exception of those involving \textit{XDER} \citep{boschini2022class}, which is not integrated into Avalanche. For \textit{XDER}, we utilize the code provided by the authors. In order to guarantee a fair comparison and in all experiments, we employ identical network architectures and maintain consistent experimental conditions.

We make certain in both settings that the buffer remains class-balanced, namely, each update maintains an equal representation of exemplars across all classes\footnote{When the buffer size is not evenly divisible by the number of classes, there may classes with an additional exemplar.}. To maintain balance, we follow this procedure: Prior to adding new exemplars we calculate $n=\frac{|\mem|}{N^t}$, where $N^t$ is the total number of existing and new classes. Subsequently, we adjust the number of exemplars from existing classes to accommodate $n$ by removing redundant points, followed by the addition of $n$ exemplars from each new class.

In the second setting, we employ a simple baseline ER-IL model. This model updates a fixed-size buffer of exemplars after training each task $t$ and replays it while training on task $t+1$. Additionally, we utilize a weighted data loader to ensure a balanced mix of data from both the new classes and the exemplars stored in the buffer in each batch. Across experiments, the only variation lies in the selection strategy employed: some experiments utilize one of the selection strategies baselines described below, and the remaining experiments employ \MethodName.

Other than this change in the method's buffer population mechanism, everything else remains the same.

\myparagraph{ER-IL baselines}\label{par:baselines}
The following ER-IL methods are 
used: 
\textit{XDER} \citep{boschini2022class}, which updates the memory buffer by integrating current and past information; \textit{ER-ACE} \citep{caccia2021new}, which applies separate losses for new and past tasks; \textit{ER} \citep{chaudhry2019continual}, which explores selection strategies in a simple ER framework; \textit{BiC} \citep{wu2019large}; \textit{iCaRL} \citep{rebuffi2017icarl}; \textit{GEM} \citep{lopez2017gradient}; and \textit{GDumb} \citep{prabhu2020gdumb}.

\begin{figure}[htbp] 
\center
\begin{subfigure}{0.1\textwidth}
\includegraphics[width=.9\linewidth]{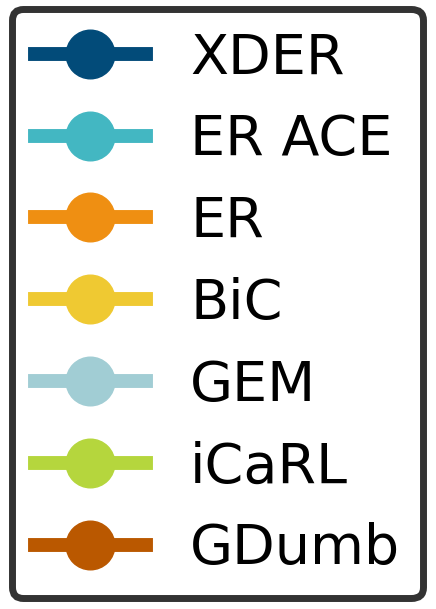}\vspace{.9cm}
\end{subfigure}
\begin{subfigure}{.325\textwidth}
\includegraphics[width=.9\linewidth]{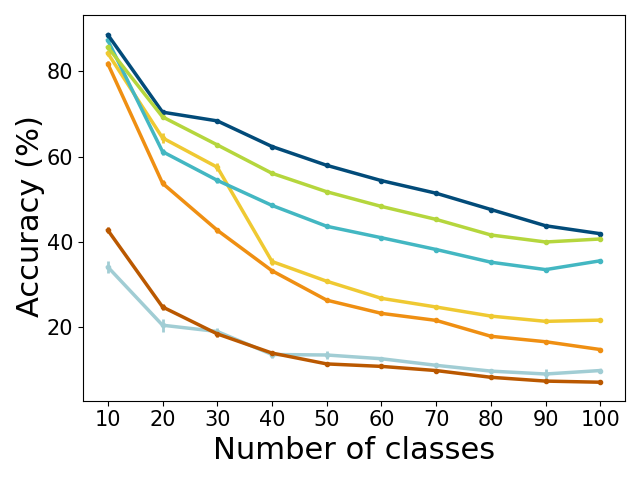}
\end{subfigure}
\caption{Baseline CIL using Split CIFAR-100 with a small buffer size $|\mem| = 300$. 7 baseline methods are shown, reporting the average accuracy $A_{t}$ as new tasks are learned. }
\label{fig:vanilla_comp}
\end{figure}

To assess the suitability of these methods under the CIL conditions studied here, we used Split CIFAR-100 and set the buffer size to 300, 500, and 2000. Figure~\ref{fig:vanilla_comp} presents the results for a buffer size of 300 (see \app\ref{app:vanila_comp} for results with larger buffer sizes). As clearly illustrated in Fig.~\ref{fig:vanilla_comp}, \textit{GEM} and \textit{GDumb} perform poorly in this scenario, likely due to their unsuitability for a very small memory buffer. Similarly, \textit{BiC} remains competitive only when the buffer size is 2000. While \textit{iCarl} remains competitive across all buffer sizes, it is inherently non-modular, relying heavily on its native \textit{Herding} selection strategy and a K-Means classifier. Hence we cannot integrate it with \MethodName.

We are therefore left with the methods \textit{XDER}, \textit{ER-ACE}, and \textit{ER}, which are both suitable and competitive for further examination with and without \MethodName\ as the selection mechanism. For \textit{ER-ACE}, which requires populating the buffer with exemplars at the start of the task before training on new classes, we cannot rely on the representation provided by a trained model. To address this, we adjust the integration of \MethodName\ by filling the buffer in two stages. First, at the beginning of task $t$, we populate the buffer with exemplars from the new classes using a random-sampling selection strategy. Second, after completing the training for task $t$, we replace the exemplars from the new classes in the buffer with those selected using \MethodName. It's important to note that throughout this process, the buffer maintains a fixed size, ensuring compliance with the class-incremental framework.

\myparagraph{Selection Strategies baselines}
\label{par:ss_baselines}
We compare with the following selection strategies (see discussion of related work in the introduction): Random sampling, \textit{Herding}, \textit{Uncertainty}, and \textit{Centered} (closest-to-center) - another common strategy that selects the exemplars closest to the center of all elements in feature space. \textit{Centered} focuses on individual exemplars, unlike \textit{Herding}, which keeps the mean of the whole constructed set close to the center. 

\myparagraph{Metrics} As customary, we employ a score that reflects the accuracy during each stage of the incremental training. Let $T$ denote the total number of tasks, and $a_{t,i}$ denote the accuracy of task $i$ at the end of task $t$ ( $i\leq t\leq T$). For every task $t={1,\dots,T}$, define its average accuracy 
$A_{t} = \frac{1}{t} \sum_{i=1}^{t}a_{t,i}$. 
This metric provides a single value for each incremental step, enabling us to directly compare different methods at each step. Since by construction each task has the same number of classes, there is no need for additional weighting terms. We do not include the metric of \textit{forgetting}, which estimates how much the model has forgotten about previous tasks, since it is less suitable for the class-incremental setting: with the addition of new classes, performance inevitably drops across all classes \citep{masana2022class}.

\myparagraph{Datasets}
We use several well known Continual Learning benchmarks. \begin{inparaenum}[(i)] \item \textbf{Split CIFAR-100} \citep{rebuffi2017icarl,DBLP:journals/corr/abs-1902-10486}, a dataset created by splitting CIFAR-100 \citep{krizhevsky2009learning} into 10 tasks, each containing 10 different classes of 32X32 images, with 500 images per class for training and 100 for testing. We also use a variation of this dataset by splitting CIFAR-100 into 20 tasks instead of 10, as reported in \app\ref{app:more_results}. \item \textbf{Split tinyImageNet}, a dataset created by splitting tinyImageNet \citep{Le2015TinyIV} into 10 tasks, each containing 20 different classes of 64X64 images, with 500 images per class for training and 50 for testing. \item \textbf{Split CUB-200} \citep{DBLP:journals/corr/abs-1812-00420}, a dataset created by splitting into 20 tasks the CUB-200 \citep{wah2011caltech} high-resolution image classification dataset, consisting of 200 categories of birds, with around 30 images per class  for training and 30 for testing. \end{inparaenum}

\myparagraph{Architectures}
\label{par:architectures}
In the first setting, except for one experiment, we employ a ResNet-18 model in all our experiments. The exception is the experiment involving the training of \textit{XDER} on the Split CUB-200 dataset. Due to computational limitations, we used instead a pre-trained ResNet-50 model \citep{DBLP:conf/cvpr/HeZRS16}, and this was maintained under all the relevant conditions. Additionally, we run some experiments with another architecture called ArchCraft, a ResNet variant designed for improved performance in CL \citep{lu2024revisiting}.
In the second setting, we use a smaller version of ResNet-18 \citep{DBLP:conf/cvpr/HeZRS16} as a simple baseline ER-IL model \citep[see][]{lopez2017gradient}.

\subsection{Main results}

\begin{figure*}[t!] 
\center
\begin{subfigure}{.7\textwidth}
\includegraphics[width=\linewidth]{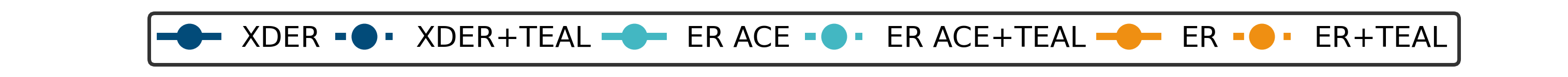}
\end{subfigure}
\\

\begin{subfigure}{.25\textwidth}
\includegraphics[width=\linewidth]{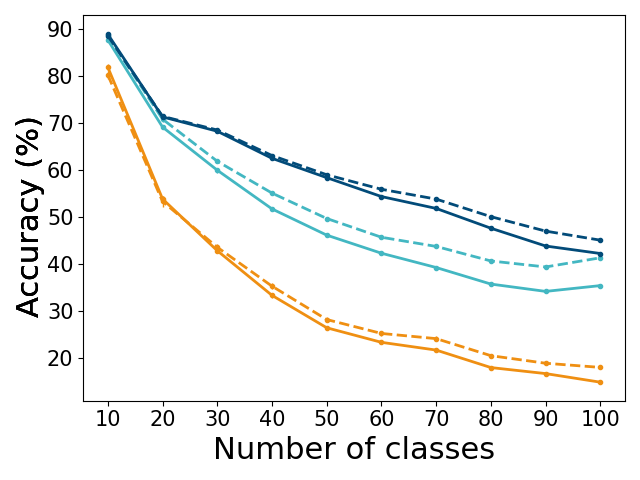}
\end{subfigure}
\hspace{.5cm}
\begin{subfigure}{.25\textwidth}
\includegraphics[width=\linewidth]{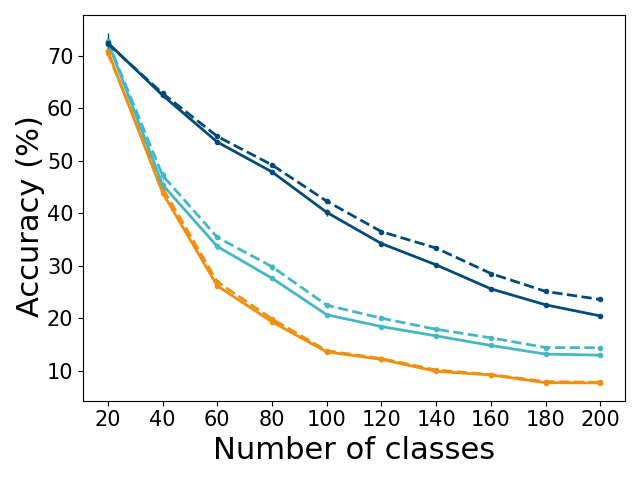}
\end{subfigure}
\hspace{.5cm}
\begin{subfigure}{.25\textwidth}
\includegraphics[width=\linewidth]{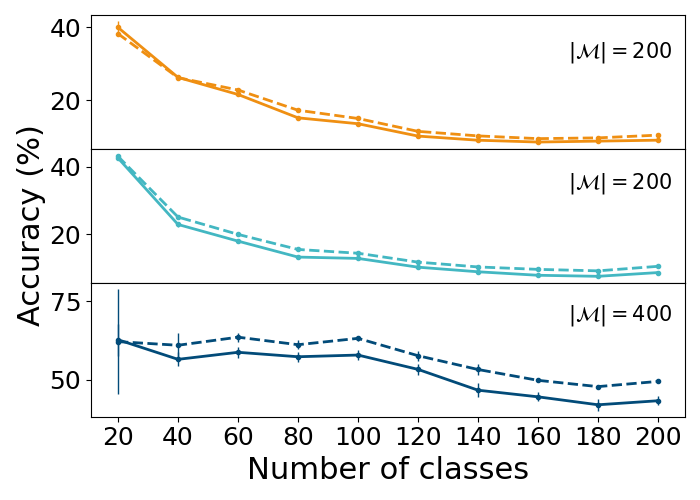}
\end{subfigure}
\\
\begin{subfigure}{.25\textwidth}
\includegraphics[width=\linewidth]{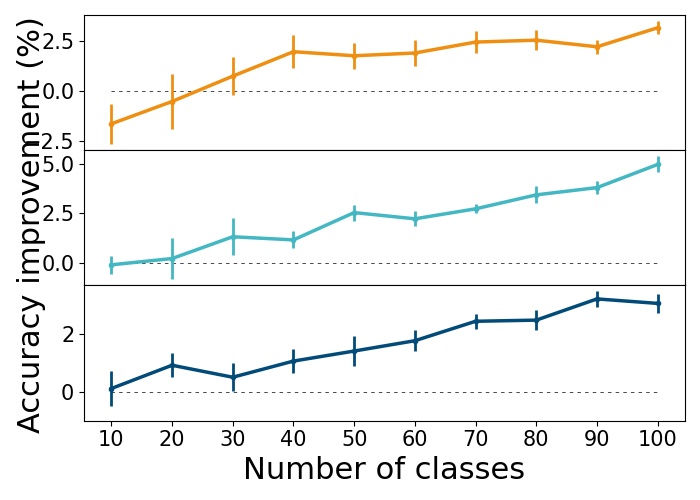}
\vspace{-0.5cm}
\caption{Split CIFAR-100, $|\mem| = 300$} 
\label{fig:accuracy_comp_a}
\end{subfigure}
\hspace{.5cm}
\begin{subfigure}{.25\textwidth}
\includegraphics[width=\linewidth]{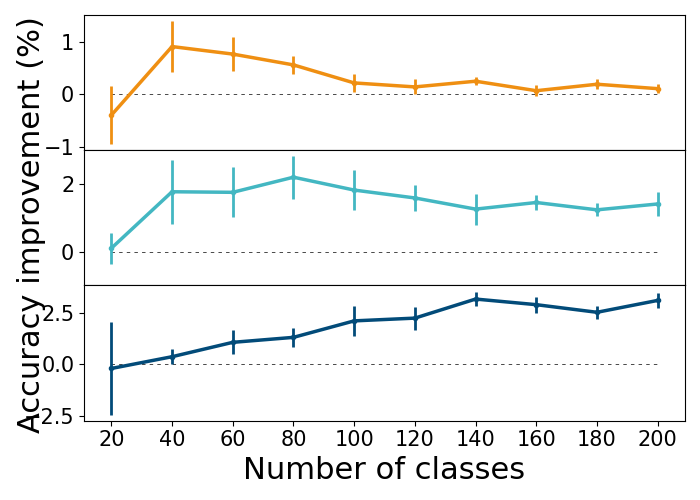}
\vspace{-0.5cm}
\caption{Split tinyImageNet, $|\mem| = 400$} 
\label{fig:accuracy_comp_b}
\end{subfigure}
\hspace{.5cm}
\begin{subfigure}{.25\textwidth}
\includegraphics[width=\linewidth]{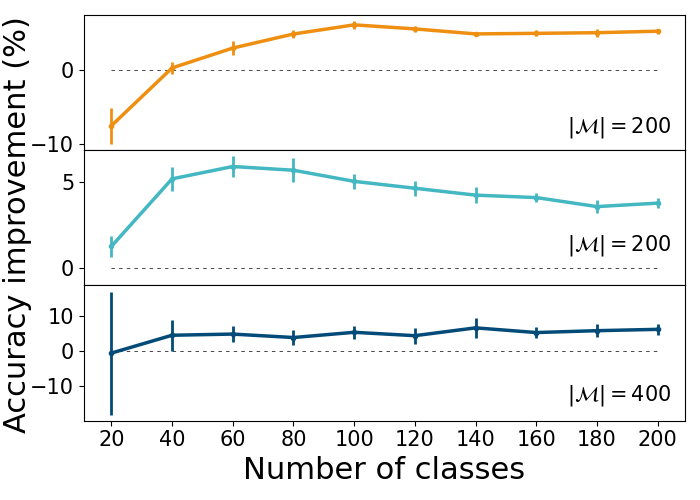}
\vspace{-0.5cm}
\caption{Split CUB-200}
\label{fig:accuracy-diff}
\end{subfigure}
\vspace{-0.2cm}
\caption{The performance of ER-IL methods with and without \MethodName. First row displays the average accuracy after training incrementally on a different number of classes. Each color corresponds to a different ER-IL method, where the continuous line represents the vanilla method, while the dashed line represents the method with \MethodName\ as its selection strategy. The error bars correspond to standard error based on 4-10 repetitions. Second row depicts the difference in accuracy between \MethodName\ and another method (\textit{XDER}, \textit{ER-ACE}, and \textit{ER}) across all tasks.
}
\label{fig:accuracy_comp}
\end{figure*}

\subsubsection{\MethodName\ integrated with SOTA ER-IL methods}
\label{sec:integrated-TEAL}

\begin{table*}[h!]
\centering
\caption{Final average accuracy for method with and without \MethodName.}
\label{tab:average_accuracy}
\small
\setlength{\tabcolsep}{4pt} 
\vspace{-0.25cm}
\begin{tabular}{llccccccccc} 
\toprule[1.2pt]
& & \multicolumn{3}{c}{XDER} & \multicolumn{3}{c}{ER-ACE} & \multicolumn{3}{c}{ER} \\
\cmidrule(lr){3-5} \cmidrule(lr){6-8} \cmidrule(lr){9-11}
Dataset & $|\mem|$ & Vanilla & +\MethodName & Improvement & Vanilla & +\MethodName & Improvement & Vanilla & +\MethodName & Improvement \\
\midrule
\multirow{3}{*}{\makecell[l]{Split\\CIFAR-100}} 
    & 100 & 27.98\tiny{±0.18} & 31.08\tiny{±0.29} & \textbf{3.1} & 24.37\tiny{±0.42} & 31.06\tiny{±0.27} & \textbf{6.9} & 10.31\tiny{±0.03} & 10.24\tiny{±0.03} & \textbf{-0.07} \\
    & 300 & 41.97\tiny{±0.25} & 45.05\tiny{±0.24} & \textbf{3.08} & 35.37\tiny{±0.29} & 41.28\tiny{±0.2} & \textbf{5.92} & 14.82\tiny{±0.21} & 17.97\tiny{±0.25} & \textbf{3.15} \\
    & 500 & 47.97\tiny{±0.22} & 50.29\tiny{±0.2} & \textbf{2.32} & 40.7\tiny{±0.25} & 45.99\tiny{±0.33} & \textbf{5.29} & 18.21\tiny{±0.29} & 22.39\tiny{±0.31} & \textbf{4.18} \\
    & 1000 & 53.69\tiny{±0.32} & 55.02\tiny{±0.34} & \textbf{1.33} & 48.16\tiny{±0.14} & 51.85\tiny{±0.22} & \textbf{3.69} & 26.66\tiny{±0.81} & 28.58\tiny{±0.65} & \textbf{1.92} \\
    & 2000 & 57.69\tiny{±0.21} & 58.79\tiny{±0.17} & \textbf{1.1} & 55.99\tiny{±0.1} & 58.7\tiny{±0.17} & \textbf{2.71} & 36.54\tiny{±0.31} & 38.17\tiny{±0.37} & \textbf{1.62} \\
    & 3000 & 58.96\tiny{±0.25} & 60.09\tiny{±0.25} & \textbf{1.13} & 59.75\tiny{±0.25} & 62.3\tiny{±0.19} & \textbf{2.55} & 43.72\tiny{±1.63} & 46.03\tiny{±1.49} & \textbf{2.32} \\
    & 4000 & 59.89\tiny{±0.16} & 60.42\tiny{±0.28} & \textbf{0.54} & 62.6\tiny{±0.22} & 64.81\tiny{±0.2} & \textbf{2.22} & 49.03\tiny{±0.89} & 52.86\tiny{±0.62} & \textbf{3.82} \\
\midrule
\multirow{3}{*}{\makecell[l]{Split\\tinyImageNet}} 
    & 200 & 14.6\tiny{±0.16} & 16.76\tiny{±0.3} & \textbf{2.16} & 11.3\tiny{±0.08} & 12.47\tiny{±0.22} & \textbf{1.17} & 7.95\tiny{±0.02} & 8.01\tiny{±0.03} & \textbf{0.07} \\
    & 400 & 20.42\tiny{±0.12} & 23.58\tiny{±0.18} & \textbf{3.16} & 13.02\tiny{±0.2} & 14.43\tiny{±0.29} & \textbf{1.41} & 7.75\tiny{±0.06} & 7.86\tiny{±0.05} & \textbf{0.11} \\
    & 600 & 25.74\tiny{±0.09} & 28.71\tiny{±0.23} & \textbf{2.97} & 14.48\tiny{±0.13} & 15.7\tiny{±0.14} & \textbf{1.22} & 7.54\tiny{±0.02} & 7.8\tiny{±0.07} & \textbf{0.26} \\
    & 1000 & 32.68\tiny{±0.24} & 34.28\tiny{±0.24} & \textbf{1.59} & 17.37\tiny{±0.34} & 18.88\tiny{±0.21} & \textbf{1.51} & 7.7\tiny{±0.06} & 7.99\tiny{±0.06} & \textbf{0.29} \\
    & 2000 & 39.76\tiny{±0.31} & 40.76\tiny{±0.23} & \textbf{1.0} & 21.71\tiny{±0.17} & 24.12\tiny{±0.24} & \textbf{2.4} & 8.27\tiny{±0.08} & 8.82\tiny{±0.09} & \textbf{0.56} \\
    & 4000 & 43.91\tiny{±0.14} & 44.42\tiny{±0.26} & \textbf{0.51} & 27.15\tiny{±0.13} & 29.62\tiny{±0.18} & \textbf{2.47} & 11.61\tiny{±0.1} & 13.59\tiny{±0.21} & \textbf{1.99} \\
    & 6000 & 44.71\tiny{±0.15} & 45.44\tiny{±0.39} & \textbf{0.73} & 30.3\tiny{±0.23} & 33.3\tiny{±0.23} & \textbf{3} & 16.53\tiny{±0.15} & 19.07\tiny{±0.18} & \textbf{2.54} \\
\midrule
\multirow{2}{*}{\makecell[l]{Split\\CUB-200}} 
    & 200 & -- & -- & \textbf{--} & 8.56\tiny{±0.15} & 10.41\tiny{±0.16} & \textbf{1.85} & 8.98\tiny{±0.18} & 10.37\tiny{±0.23} & \textbf{1.39} \\
    & 400 & 43.63\tiny{±1.78} & 49.55\tiny{±0.47} & \textbf{5.96} & 11.25\tiny{±0.99} & 12.33\tiny{±0.26} & \textbf{1.08} & 12.01\tiny{±0.52} & 14.13\tiny{±0.3} & \textbf{2.12} \\
\bottomrule[1.2pt]
\end{tabular}
\end{table*}

As mentioned above, we investigate the baseline methods \textit{XDER}, \textit{ER-ACE}, and \textit{ER}, comparing their performance using either their native selection strategies or \MethodName. We conducted experiments on Split CIFAR-100 with buffer sizes ranging from 100 to 4000, Split tinyImageNet with buffer sizes from 200 to 6000, and Split CUB-200 with buffer sizes of 200\footnote{Due to computational limitations of the relevant package, we did not run \textit{XDER} with a buffer size of 200.} and 400. The results for final average accuracy $A_{t}$ and the differences between 'improved' and 'vanilla' are presented in Table~\ref{tab:average_accuracy}. 
First row of Fig.~\ref{fig:accuracy_comp} shows the average accuracy $A_{t}$ after each task $t$, while the second row illustrates the corresponding improvement from \MethodName\ in selected experiments.

As demonstrated in Fig.~\ref{fig:accuracy_comp}, in most cases, the improvement achieved by \MethodName\ increases as incremental training progresses, resulting in a more significant improvement by the final task. Indeed, we do not expect significant differences immediately after the first task, as catastrophic forgetting hasn't occurred yet. Reassuringly, for the majority of the experiments, the improvement becomes apparent starting from the second task.

Table \ref{tab:average_accuracy} shows that the difference (indicated in the 'improvement' row) is consistently positive, demonstrating that \MethodName\ always enhances performance. However, it is important to note that when catastrophic forgetting in the vanilla version is too severe, the integration of \MethodName\ can no longer mend the damage. Consequently, the enhancements to \textit{ER} and \textit{ER-ACE} on the Split tinyImageNet dataset are mostly notable with larger buffer sizes. 

\begin{figure}[H]
    \center
    \begin{subfigure}{.8\columnwidth}
        \includegraphics[width=\linewidth]{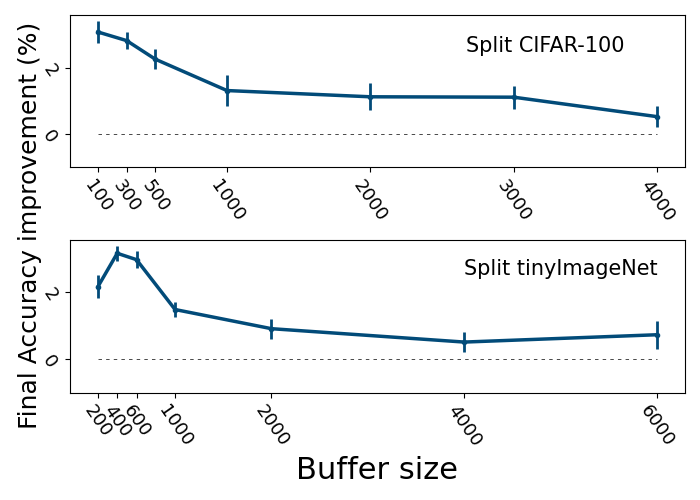}
    \end{subfigure}
\caption{Performance improvement of \MethodName\ when integrated with \textit{XDER} over various buffer sizes.
}
\label{fig:xder_buffer_comp}
\end{figure}

Note that the improvement provided by \MethodName\ tends to increase as the buffer size decreases. To further explore the relationship between buffer size and relative improvement, we present a graph showing the final average accuracy improvement across various buffer sizes. The results for Split CIFAR-100 and Split tinyImageNet using \textit{XDER} are illustrated in Fig.~\ref{fig:xder_buffer_comp} (see results with \textit{ER-ACE} and \textit{ER} in \app\ref{app:buffer_comp}). Clearly, while improvements do occur across all buffer sizes,  larger improvements are seen for smaller buffers.

\subsubsection{Comparison of selection strategy} 
\label{sec:herding-TEAL}
\begin{figure}[htb]
    \center
    \begin{subfigure}{\textwidth}
        \includegraphics[width=0.45\linewidth]{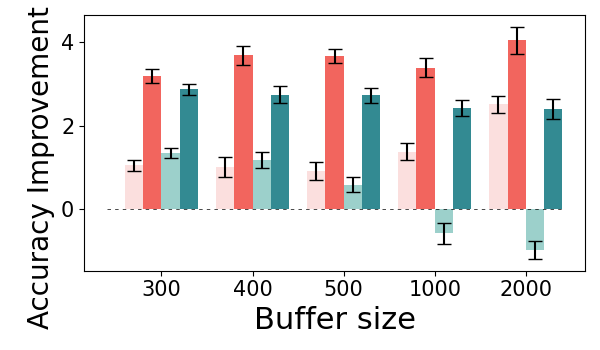}
    \end{subfigure}
\hfill
    \begin{subfigure}{\textwidth}
        \scriptsize
        \hspace{1.3cm} \includegraphics[width=0.35\linewidth]{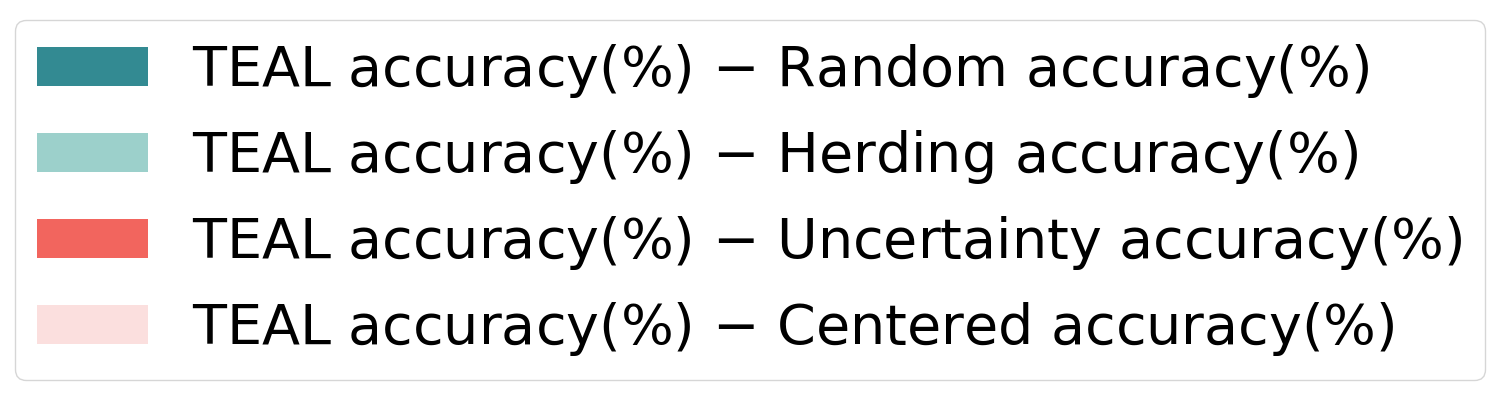}
    \end{subfigure}
\vspace{-0.4cm}
\caption{Split CIFAR-100: \MethodName\ performance gain compared to 4 baselines: random-sampling, \textit{Herding}, \textit{Centered}, and \textit{Uncertainty}, each bar corresponds to a fixed buffer size. The error bars correspond to standard error based on 10-20 repetitions. 
}
\vspace{-0.2cm}
\label{fig:selection_strategies_comp}
\end{figure}

We begin with a simple ER-IL model on Split CIFAR-100\footnote{The order of the classes and the partition to tasks is randomly selected and fixed in all conditions.} with various buffer sizes, and 5 different selection strategies: random, \textit{Herding}, \textit{Uncertainty}, \textit{Centered} and \MethodName.
As depicted in Fig.~\ref{fig:selection_strategies_comp}, our method enhances performance as compared to \textit{Uncertainty} by 3-4\%, and random selection by almost 3\% for smaller buffers (300, 400, 500) and about 2.5\% for larger buffers (1000, 2000). When compared to \textit{Herding}, our method is more effective with smaller buffer sizes, enhancing performance by up to 1.2\% for a buffer size of 300, but less effective with large buffers. It is possible that the mechanism of retaining a set of the nearest neighbors around the average sample in each class becomes less effective when the buffer is too small. Additionally, \textit{Herding} lacks a component for selecting a diverse set, which may lead to the retention of similar exemplars, an issue that becomes more pronounced with smaller buffer sizes. Interestingly, despite its similarity to \textit{Herding}, \textit{Centered} performs worse on larger buffer sizes, resulting in a 1.5-2.5\% improvement by \MethodName.
Similar results, using different class ordering, are shown in \app\ref{app:stand-alone}.

\begin{figure*}[htbp] 
\center
\begin{subfigure}{.3\textwidth}
\includegraphics[width=\linewidth]{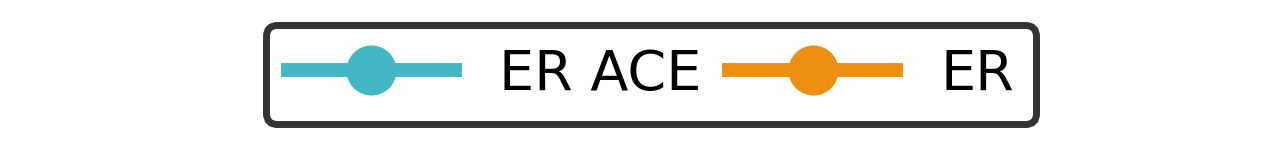}
\end{subfigure}

\begin{subfigure}{.225\textwidth}
\includegraphics[width=\linewidth]{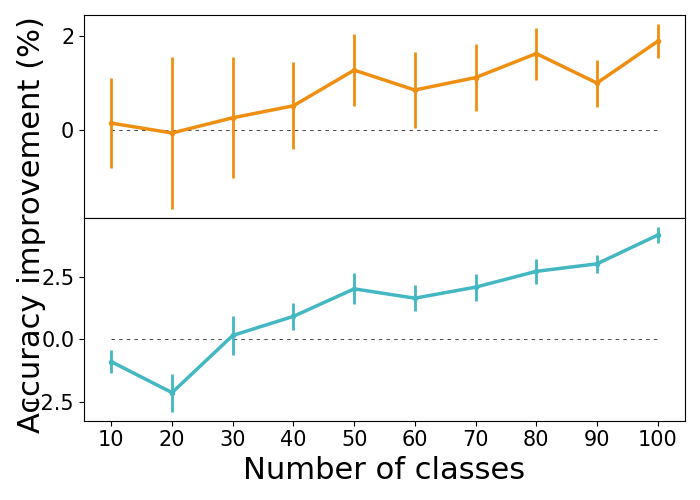}
\caption{$|\mem| = 300$} 
\end{subfigure}
\begin{subfigure}{.225\textwidth}
\includegraphics[width=\linewidth]{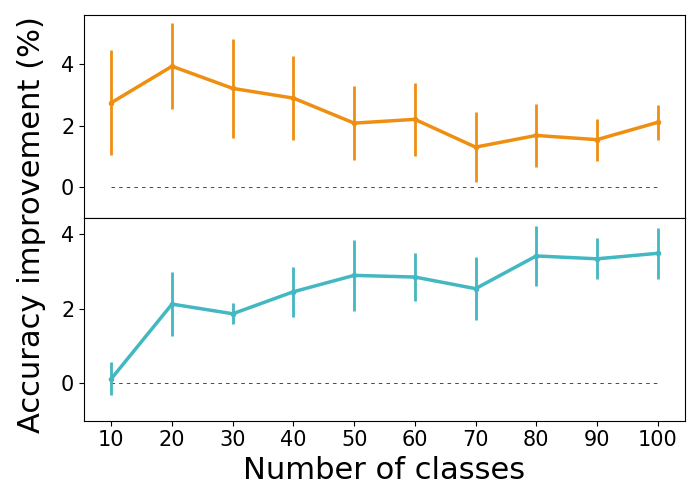}
\caption{$|\mem| = 500$} 
\end{subfigure}
\begin{subfigure}{.225\textwidth}
\includegraphics[width=\linewidth]{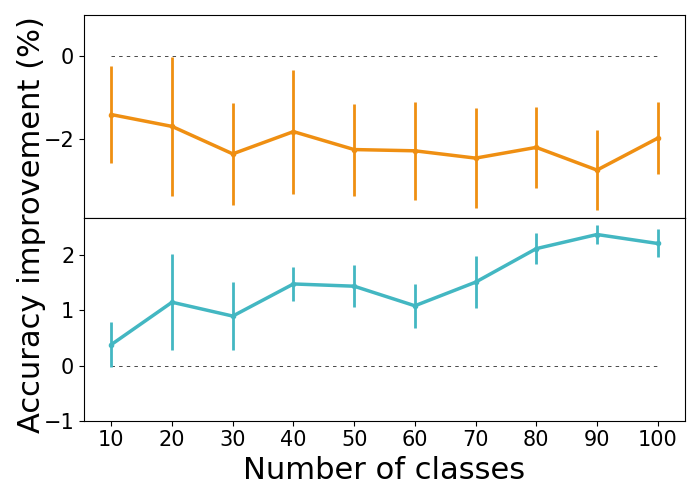}
\caption{$|\mem| = 2000$}
\end{subfigure}

\caption{The difference between the improvement obtained by \MethodName\ and the one obtained by \textit{Herding}, showing 2 ER-IL methods and 3 buffer sizes while training on the Split CIFAR-100 dataset. 
}
\label{fig:herding_comp}
\vspace{-0.3cm}
\end{figure*}

Given the results shown in Fig.~\ref{fig:selection_strategies_comp}, 
we repeat this comparison while integrating both methods - \MethodName\ and \textit{Herding} - into competitive ER-IL methods.  Fig.~\ref{fig:herding_comp} shows the difference between the improvement obtained by \MethodName\ and the one obtained by \textit{Herding}. Since in almost all cases this difference is positive, we may conclude that \MethodName\ is significantly more beneficial than \textit{Herding}.

\subsubsection{Results of Task-IL}
\label{sec:task_il}
The incorporation of \MethodName\ into existing ER-IL method is also beneficial in the task incremental scenario, although the advantage is less pronounced, likely because the scenario is easier. Results are shown in Table~\ref{tab:task_il_average_accuracy}.

\begin{table}[h!]
\centering
\caption{\textbf{Task-IL}, accuracy with and without \MethodName} 
\label{tab:task_il_average_accuracy}
\small
\setlength{\tabcolsep}{3.7pt} 
\vspace{-0.25cm}
\begin{tabular}{lcccccc} 
\toprule[1.2pt]
Dataset & $|\mem|$ & XDER & XDER+\MethodName & Improvement \\ 
\midrule
\multirow{3}{*}{\shortstack{Split\\CIFAR-100}} & 100 & 73.75\tiny{±0.59} & 76.56\tiny{±0.27} & \textbf{2.81} \\
                                 & 300 & 83.5\tiny{±0.06} & 84.46\tiny{±0.26} & \textbf{0.96} \\
                                 & 500 & 85.96\tiny{±0.14} & 86.1\tiny{±0.2} & \textbf{0.14} \\ 
\midrule
\multirow{4}{*}{\shortstack{Split\\tinyImageNet}} & 200 & 42.48\tiny{±0.38} & 46.55\tiny{±0.37} & \textbf{4.07} \\
                                    & 400 & 55.03\tiny{±0.15} & 59.2\tiny{±0.47} & \textbf{4.17} \\
                                    & 600 & 63.69\tiny{±0.21} & 65.26\tiny{±0.28} & \textbf{1.57} \\
                                    & 1000 & 69.72\tiny{±0.31} & 70.61\tiny{±0.23} & \textbf{0.89} \\
\bottomrule[1.2pt]
\end{tabular}
\end{table}

\subsection{Ablation study}
\label{sec:ablation}

\subsubsection{Iterative process of \MethodName}
\label{sec:ablation_iterative}
We explore an alternative variant of \MethodName, which selects the set of exemplars in a single pass instead of by iterations. We call this variant \MethodName\_\textit{OneTime}. While still selecting exemplars for each class separately, this variant first partitions the exemplars of each class into $n=\frac{|\mem|}{N^t}$ clusters, and then selects the most typical exemplar from each cluster in descending order of cluster size, preserving the order of selections.

\begin{table}[h]
\centering
\caption{The improvement of 2 variants of \MethodName\ when integrated with different CIL methods: \textit{XDER}, \textit{ER-ACE}, and \textit{ER}. In almost all cases, the original \MethodName\ matches or outperforms the \textit{OneTime} variant.
\label{tab:ablation}}
\small
\setlength{\tabcolsep}{5.3pt} 
\begin{tabular}{llccc} 
\toprule[1.2pt]
$|\mem|$ & Selection Strategy & XDER & ER-ACE & ER \\ 
\midrule
\multirow{2}{*}{300} & \MethodName\_\emph{OneTime} & 44.13\tiny{±0.26} & 35.81\tiny{±0.42} & 17.64\tiny{±0.12} \\
& \MethodName & \textbf{45.05\tiny{±0.24}} & \textbf{36.4\tiny{±0.12}} & \textbf{18.1\tiny{±0.1}} \\
\midrule
\multirow{2}{*}{500} & \MethodName\_\emph{OneTime} & 49.87\tiny{±0.22} & 40.16\tiny{±0.3} & 22.01\tiny{±0.18} \\
& \MethodName & \textbf{50.29\tiny{±0.2}} & 40.26\tiny{±0.23} & 22.07\tiny{±0.13} \\
\midrule
\multirow{2}{*}{2000} & \MethodName\_\emph{OneTime} & 59.05\tiny{±0.14} & 51.52\tiny{±0.24} & \textbf{40.65\tiny{±0.32}} \\
& \MethodName & 58.79\tiny{±0.17} & 51.68\tiny{±0.35} & 39.96\tiny{±0.23} \\
\bottomrule[1.2pt]
\end{tabular}
\end{table}

\begin{table}[h!]
\scriptsize
\centering
\caption{Final average accuracy for method with and without \MethodName\ using the ArchCraft architecture. Top 3 rows correspond to Split CIFAR-100, bottom 3 to Split tinyImageNet.}
\label{tab:arch_craft}
\small
\setlength{\tabcolsep}{2.8pt} 
\begin{tabular}{ccccccc} 
\toprule[1.2pt]
& \multicolumn{3}{c}{ER-ACE} & \multicolumn{3}{c}{ER} \\
\cmidrule(lr){2-4} \cmidrule(lr){5-7}
$|\mem|$ & Vanilla & +\MethodName & Imp. & Vanilla & +\MethodName & Imp. \\
\midrule
300 & 36.75\tiny{±0.28} & 42.47\tiny{±0.08} & \textbf{5.72} & 14.59\tiny{±0.08} & 18.01\tiny{±0.35} & \textbf{3.42} \\
   500 & 41.98\tiny{±0.24} & 47.53\tiny{±0.35} & \textbf{5.56} & 18.54\tiny{±0.4} & 22.3\tiny{±0.4} & \textbf{3.76} \\
   2000 & 57.33\tiny{±0.18} & 60.18\tiny{±0.17} & \textbf{2.85} & 41.03\tiny{±0.37} & 43.44\tiny{±0.28} & \textbf{2.41} \\
\midrule
400 & 12.24\tiny{±0.11} & 13.2\tiny{±0.38} & \textbf{0.96} & 7.67\tiny{±0.13} & 7.84\tiny{±0.19} & \textbf{0.17} \\
   1000 & 15.64\tiny{±0.37} & 16.7\tiny{±0.2} & \textbf{1.07} & 8.12\tiny{±0.05} & 8.36\tiny{±0.18} & \textbf{0.24} \\
   2000 & 19.08\tiny{±0.36} & 20.43\tiny{±0.37} & \textbf{1.35} & 9.65\tiny{±0.11} & 10.66\tiny{±0.03} & \textbf{1.01} \\
\bottomrule[1.2pt]
\end{tabular}
\end{table}

The experiments on the two variants are conducted using Split CIFAR-100 with the smaller version of ResNet-18 mentioned in Section~\ref{par:architectures} and buffer sizes of 300, 500 and 2000. We investigate the performance enhancement of each variant when integrated with \textit{XDER}, \textit{ER-ACE}, and \textit{ER}. The results are shown in Table~\ref{tab:ablation}. Clearly, whenever there is a significant difference in performance, the iterative variant outperforms the other one in 4 out of 5 cases. 

\subsubsection{Different architecture}
\label{sec:ablation_arch}
Some experiments were replicated using ArchCraft instead of the ResNet-18 architecture, obtaining similar results (see Table~\ref{tab:arch_craft}).

\section{Summary and discussion}

We proposed a new mechanism to select exemplars for the memory buffer in replay-based CIL methods. This method is based on the principles of diversity and representativeness. When the memory buffer is relatively small, our method  \MethodName\ is shown to outperform both the native mechanism of each ER-IL method (usually random selection) and alternative selection mechanisms, including \textit{Herding}, \textit{Uncertainty} and \textit{Centered}. Even when the buffer is large, our method is beneficial in almost all cases. 

In future work we will investigate ways to determine at which point, if any, the added value of our method diminishes or possibly becomes harmful. We will also integrate \MethodName\ with other replay-based CIL methods that utilize different selection strategies.

\clearpage
\bibliography{aaai25}

\begin{thebibliography}{34}
\providecommand{\natexlab}[1]{#1}

\bibitem[{Aljundi et~al.(2019)Aljundi, Lin, Goujaud, and
  Bengio}]{aljundi2019gradient}
Aljundi, R.; Lin, M.; Goujaud, B.; and Bengio, Y. 2019.
\newblock Gradient based sample selection for online continual learning.
\newblock \emph{Advances in neural information processing systems}, 32.

\bibitem[{Bang et~al.(2021)Bang, Kim, Yoo, Ha, and Choi}]{bang2021rainbow}
Bang, J.; Kim, H.; Yoo, Y.; Ha, J.-W.; and Choi, J. 2021.
\newblock Rainbow memory: Continual learning with a memory of diverse samples.
\newblock In \emph{Proceedings of the IEEE/CVF conference on computer vision
  and pattern recognition}, 8218--8227.

\bibitem[{Boschini et~al.(2022)Boschini, Bonicelli, Buzzega, Porrello, and
  Calderara}]{boschini2022class}
Boschini, M.; Bonicelli, L.; Buzzega, P.; Porrello, A.; and Calderara, S. 2022.
\newblock Class-incremental continual learning into the extended der-verse.
\newblock \emph{IEEE transactions on pattern analysis and machine
  intelligence}, 45(5): 5497--5512.

\bibitem[{Caccia et~al.(2021)Caccia, Aljundi, Asadi, Tuytelaars, Pineau, and
  Belilovsky}]{caccia2021new}
Caccia, L.; Aljundi, R.; Asadi, N.; Tuytelaars, T.; Pineau, J.; and Belilovsky,
  E. 2021.
\newblock New insights on reducing abrupt representation change in online
  continual learning.
\newblock \emph{arXiv preprint arXiv:2104.05025}.

\bibitem[{Chaudhry et~al.(2018{\natexlab{a}})Chaudhry, Dokania, Ajanthan, and
  Torr}]{chaudhry2018riemannian}
Chaudhry, A.; Dokania, P.~K.; Ajanthan, T.; and Torr, P.~H. 2018{\natexlab{a}}.
\newblock Riemannian walk for incremental learning: Understanding forgetting
  and intransigence.
\newblock In \emph{Proceedings of the European conference on computer vision
  (ECCV)}, 532--547.

\bibitem[{Chaudhry et~al.(2018{\natexlab{b}})Chaudhry, Ranzato, Rohrbach, and
  Elhoseiny}]{DBLP:journals/corr/abs-1812-00420}
Chaudhry, A.; Ranzato, M.; Rohrbach, M.; and Elhoseiny, M. 2018{\natexlab{b}}.
\newblock Efficient Lifelong Learning with {A-GEM}.
\newblock \emph{CoRR}, abs/1812.00420.

\bibitem[{Chaudhry et~al.(2019{\natexlab{a}})Chaudhry, Rohrbach, Elhoseiny,
  Ajanthan, Dokania, Torr, and Ranzato}]{chaudhry2019continual}
Chaudhry, A.; Rohrbach, M.; Elhoseiny, M.; Ajanthan, T.; Dokania, P.; Torr, P.;
  and Ranzato, M. 2019{\natexlab{a}}.
\newblock Continual learning with tiny episodic memories.
\newblock In \emph{Workshop on Multi-Task and Lifelong Reinforcement Learning}.

\bibitem[{Chaudhry et~al.(2019{\natexlab{b}})Chaudhry, Rohrbach, Elhoseiny,
  Ajanthan, Dokania, Torr, and Ranzato}]{DBLP:journals/corr/abs-1902-10486}
Chaudhry, A.; Rohrbach, M.; Elhoseiny, M.; Ajanthan, T.; Dokania, P.~K.; Torr,
  P. H.~S.; and Ranzato, M. 2019{\natexlab{b}}.
\newblock Continual Learning with Tiny Episodic Memories.
\newblock \emph{CoRR}, abs/1902.10486.

\bibitem[{Choi, El-Khamy, and Lee(2021)}]{choi2021dual}
Choi, Y.; El-Khamy, M.; and Lee, J. 2021.
\newblock Dual-teacher class-incremental learning with data-free generative
  replay.
\newblock In \emph{Proceedings of the IEEE/CVF Conference on Computer Vision
  and Pattern Recognition}, 3543--3552.

\bibitem[{De~Lange et~al.(2021)De~Lange, Aljundi, Masana, Parisot, Jia,
  Leonardis, Slabaugh, and Tuytelaars}]{de2021continual}
De~Lange, M.; Aljundi, R.; Masana, M.; Parisot, S.; Jia, X.; Leonardis, A.;
  Slabaugh, G.; and Tuytelaars, T. 2021.
\newblock A continual learning survey: Defying forgetting in classification
  tasks.
\newblock \emph{IEEE transactions on pattern analysis and machine
  intelligence}, 44(7): 3366--3385.

\bibitem[{Gao and Liu(2023)}]{gao2023ddgr}
Gao, R.; and Liu, W. 2023.
\newblock Ddgr: Continual learning with deep diffusion-based generative replay.
\newblock In \emph{International Conference on Machine Learning}, 10744--10763.
  PMLR.

\bibitem[{Gautam et~al.(2024)Gautam, Parameswaran, Mishra, and
  Sundaram}]{gautam2024generative}
Gautam, C.; Parameswaran, S.; Mishra, A.; and Sundaram, S. 2024.
\newblock Generative replay-based continual zero-shot learning.
\newblock In \emph{Towards Human Brain Inspired Lifelong Learning}, 73--100.
  World Scientific.

\bibitem[{Hacohen, Dekel, and Weinshall(2022)}]{hacohen2022active}
Hacohen, G.; Dekel, A.; and Weinshall, D. 2022.
\newblock Active Learning on a Budget: Opposite Strategies Suit High and Low
  Budgets.
\newblock In \emph{International Conference on Machine Learning}. PMLR.

\bibitem[{Hacohen and Tuytelaars(2024)}]{hacohen2024forgetting}
Hacohen, G.; and Tuytelaars, T. 2024.
\newblock Forgetting Order of Continual Learning: Examples That are Learned
  First are Forgotten Last.
\newblock \emph{arXiv preprint arXiv:2406.09935}.

\bibitem[{He et~al.(2016)He, Zhang, Ren, and Sun}]{DBLP:conf/cvpr/HeZRS16}
He, K.; Zhang, X.; Ren, S.; and Sun, J. 2016.
\newblock Deep Residual Learning for Image Recognition.
\newblock In \emph{2016 {IEEE} Conference on Computer Vision and Pattern
  Recognition, {CVPR} 2016, Las Vegas, NV, USA, June 27-30, 2016}, 770--778.
  {IEEE} Computer Society.

\bibitem[{Krizhevsky, Hinton et~al.(2009)}]{krizhevsky2009learning}
Krizhevsky, A.; Hinton, G.; et~al. 2009.
\newblock Learning multiple layers of features from tiny images.
\newblock \emph{Online}.

\bibitem[{Le and Yang(2015)}]{Le2015TinyIV}
Le, Y.; and Yang, X.~S. 2015.
\newblock Tiny ImageNet Visual Recognition Challenge.

\bibitem[{Li and Hoiem(2017)}]{li2017learning}
Li, Z.; and Hoiem, D. 2017.
\newblock Learning without forgetting.
\newblock \emph{IEEE transactions on pattern analysis and machine
  intelligence}, 40(12): 2935--2947.

\bibitem[{Lloyd(1982)}]{lloyd1982least}
Lloyd, S. 1982.
\newblock Least squares quantization in PCM.
\newblock \emph{IEEE transactions on information theory}, 28(2): 129--137.

\bibitem[{Lomonaco et~al.(2021)Lomonaco, Pellegrini, Cossu, Carta, Graffieti,
  Hayes, De~Lange, Masana, Pomponi, Van~de Ven et~al.}]{lomonaco2021avalanche}
Lomonaco, V.; Pellegrini, L.; Cossu, A.; Carta, A.; Graffieti, G.; Hayes,
  T.~L.; De~Lange, M.; Masana, M.; Pomponi, J.; Van~de Ven, G.~M.; et~al. 2021.
\newblock Avalanche: an end-to-end library for continual learning.
\newblock In \emph{Proceedings of the IEEE/CVF Conference on Computer Vision
  and Pattern Recognition}, 3600--3610.

\bibitem[{Lopez-Paz and Ranzato(2017)}]{lopez2017gradient}
Lopez-Paz, D.; and Ranzato, M. 2017.
\newblock Gradient episodic memory for continual learning.
\newblock \emph{Advances in neural information processing systems}, 30.

\bibitem[{Lu et~al.(2024)Lu, Feng, Yuan, Song, and Sun}]{lu2024revisiting}
Lu, A.; Feng, T.; Yuan, H.; Song, X.; and Sun, Y. 2024.
\newblock Revisiting Neural Networks for Continual Learning: An Architectural
  Perspective.
\newblock \emph{arXiv preprint arXiv:2404.14829}.

\bibitem[{Mallya and Lazebnik(2018)}]{mallya2018packnet}
Mallya, A.; and Lazebnik, S. 2018.
\newblock Packnet: Adding multiple tasks to a single network by iterative
  pruning.
\newblock In \emph{Proceedings of the IEEE conference on Computer Vision and
  Pattern Recognition}, 7765--7773.

\bibitem[{Masana et~al.(2022)Masana, Liu, Twardowski, Menta, Bagdanov, and Van
  De~Weijer}]{masana2022class}
Masana, M.; Liu, X.; Twardowski, B.; Menta, M.; Bagdanov, A.~D.; and Van
  De~Weijer, J. 2022.
\newblock Class-incremental learning: survey and performance evaluation on
  image classification.
\newblock \emph{IEEE Transactions on Pattern Analysis and Machine
  Intelligence}, 45(5): 5513--5533.

\bibitem[{McCloskey and Cohen(1989)}]{mccloskey1989catastrophic}
McCloskey, M.; and Cohen, N.~J. 1989.
\newblock Catastrophic interference in connectionist networks: The sequential
  learning problem.
\newblock In \emph{Psychology of learning and motivation}, volume~24, 109--165.
  Elsevier.

\bibitem[{Prabhu et~al.(2023)Prabhu, Al~Kader~Hammoud, Dokania, Torr, Lim,
  Ghanem, and Bibi}]{prabhu2023computationally}
Prabhu, A.; Al~Kader~Hammoud, H.~A.; Dokania, P.~K.; Torr, P.~H.; Lim, S.-N.;
  Ghanem, B.; and Bibi, A. 2023.
\newblock Computationally budgeted continual learning: What does matter?
\newblock In \emph{Proceedings of the IEEE/CVF Conference on Computer Vision
  and Pattern Recognition}, 3698--3707.

\bibitem[{Prabhu, Torr, and Dokania(2020)}]{prabhu2020gdumb}
Prabhu, A.; Torr, P.~H.; and Dokania, P.~K. 2020.
\newblock Gdumb: A simple approach that questions our progress in continual
  learning.
\newblock In \emph{Computer Vision--ECCV 2020: 16th European Conference,
  Glasgow, UK, August 23--28, 2020, Proceedings, Part II 16}, 524--540.
  Springer.

\bibitem[{Rebuffi et~al.(2017)Rebuffi, Kolesnikov, Sperl, and
  Lampert}]{rebuffi2017icarl}
Rebuffi, S.-A.; Kolesnikov, A.; Sperl, G.; and Lampert, C.~H. 2017.
\newblock icarl: Incremental classifier and representation learning.
\newblock In \emph{Proceedings of the IEEE conference on Computer Vision and
  Pattern Recognition}, 2001--2010.

\bibitem[{Shin et~al.(2017)Shin, Lee, Kim, and Kim}]{shin2017continual}
Shin, H.; Lee, J.~K.; Kim, J.; and Kim, J. 2017.
\newblock Continual learning with deep generative replay.
\newblock \emph{Advances in neural information processing systems}, 30.

\bibitem[{Tian et~al.(2024)Tian, Li, Li, Ran, Ning, and
  Tiwari}]{tian2024survey}
Tian, S.; Li, L.; Li, W.; Ran, H.; Ning, X.; and Tiwari, P. 2024.
\newblock A survey on few-shot class-incremental learning.
\newblock \emph{Neural Networks}, 169: 307--324.

\bibitem[{Van~de Ven, Tuytelaars, and Tolias(2022)}]{van2022three}
Van~de Ven, G.~M.; Tuytelaars, T.; and Tolias, A.~S. 2022.
\newblock Three types of incremental learning.
\newblock \emph{Nature Machine Intelligence}, 4(12): 1185--1197.

\bibitem[{Wah et~al.(2011)Wah, Branson, Welinder, Perona, and
  Belongie}]{wah2011caltech}
Wah, C.; Branson, S.; Welinder, P.; Perona, P.; and Belongie, S. 2011.
\newblock The caltech-ucsd birds-200-2011 dataset.

\bibitem[{Welling(2009)}]{welling2009herding}
Welling, M. 2009.
\newblock Herding dynamical weights to learn.
\newblock In \emph{Proceedings of the 26th annual international conference on
  machine learning}, 1121--1128.

\bibitem[{Wu et~al.(2019)Wu, Chen, Wang, Ye, Liu, Guo, and Fu}]{wu2019large}
Wu, Y.; Chen, Y.; Wang, L.; Ye, Y.; Liu, Z.; Guo, Y.; and Fu, Y. 2019.
\newblock Large scale incremental learning.
\newblock In \emph{Proceedings of the IEEE/CVF conference on computer vision
  and pattern recognition}, 374--382.

\end{thebibliography}
\clearpage

\appendix
\section*{Supplementary Material}

\section{Implementation details}
\label{app:impl_details}

The source code for this study can be found in the zip file included in the supplementary material. This zip file contains a README.md file that provides instructions on how to run the experiments. The code would be uploaded to our GitHub repository and will be made public upon acceptance.

\subsection{\MethodName\ implementation}

\paragraph{Clustering algorithm} We used scikit-learn KMeans implementation.

\paragraph{Iterations pace} \MethodName, as described in Alg. \ref{alg:construct_exemplar_set}, requires an iterations pace $s_1\leq\dots\leq s_k=n$ which indicates the pace of selecting exemplars from new class data. In both settings we use a logarithmic pace. We set a base $b=1.4$, and define $s_1=\floor{b^4}, s_2=\floor{b^5},\dots, s_k=\floor{b^{log_{b}n}}$.

\subsection{Stand-alone setting}
\label{app:stand-alone}

For all selection strategies, we use a smaller ResNet-18, as mentioned above, trained for 200 epochs. Optimization is performed with an SGD optimizer using Nesterov momentum of 0.9, a weight decay of 0.0002, and a learning rate that starts at 0.1 and decays by a factor of 0.3 every 66 epochs. Training is conducted with a batch size of 128 examples, and data augmentation is applied through random cropping and horizontal flips.

\begin{figure}[h] 
\center
\begin{subfigure}{.23\textwidth}
\includegraphics[width=\linewidth]{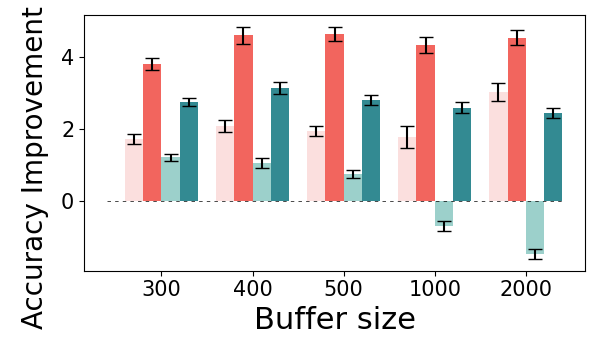}
\caption{Order 2}
\label{fig:selection_strategy_2}
\end{subfigure}
\begin{subfigure}{.23\textwidth}
\includegraphics[width=\linewidth]{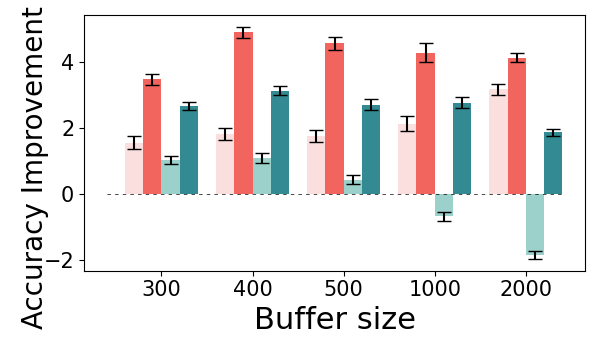}
\caption{Order 3}
\label{fig:selection_strategy_3}
\end{subfigure}
\begin{subfigure}{.35\textwidth}
\includegraphics[width=\linewidth]{figures/cifar100/selection_strategires/legends_diff_all.png}
\end{subfigure}
\vspace{-0.3cm}
\caption{Split CIFAR-100: \MethodName\ Enhanced accuracy of \MethodName\ compared to 4 baselines: random-sampling, \textit{Herding}, \textit{Centered}, and \textit{Uncertainty},  in two different classes order.}
\label{fig:app_selection_strategies_comp}
\vspace{-0.5cm}
\end{figure}

We run this setting on Split CIFAR-100 with three different random orders of classes:

\begin{quote}
\hspace{-.5cm} 1.
The order we display on Fig.~\ref{fig:selection_strategies_comp} is: [44, 19, 93, 90, 71, 69, 37, 95, 53, 91, 81, 42, 80, 85, 74, 56, 76, 63, 82, 40, 26, 92, 57, 10, 16, 66, 89, 41, 97, 8, 31, 24, 35, 30, 65, 7, 98, 23, 20, 29, 78, 61, 94, 15, 4, 52, 59, 5, 54, 46, 3, 28, 2, 70, 6, 60, 49, 68, 55, 72, 79, 77, 45, 1, 32, 34, 11, 0, 22, 12, 87, 50, 25, 47, 36, 96, 9, 83, 62, 84, 18, 17, 75, 67, 13, 48, 39, 21, 64, 88, 38, 27, 14, 73, 33, 58, 86, 43, 99, 51] \\
\vspace{-.4cm}
\end{quote}
\begin{quote}
\hspace{-.5cm} 2.
    The order in Fig.~\ref{fig:selection_strategy_2} is: [45, 15, 90, 32, 35, 63, 17, 72, 79, 96, 48, 36, 16, 11, 23, 80, 22, 58, 3, 62, 50, 33, 66, 99, 43, 76, 7, 57, 81, 82, 6, 10, 24, 52, 95, 73, 91, 21, 38, 31, 85, 59, 13, 69, 75, 70, 64, 8, 77, 34, 46, 39, 92, 0, 44, 98, 49, 9, 4, 61, 12, 83, 28, 78, 40, 88, 54, 5, 26, 41, 89, 20, 84, 2, 1, 55, 19, 74, 25, 37, 42, 14, 30, 18, 67, 71, 68, 27, 60, 51, 29, 56, 93, 47, 97, 94, 86, 87, 65, 53]\\
\end{quote}
\vspace{-.3cm}
\begin{quote}
\hspace{-.4cm} 3.
    The order in Fig.~\ref{fig:selection_strategy_3} is: [48, 97, 1, 81, 90, 49, 10, 8, 7, 20, 70, 73, 75, 14, 91, 38, 47, 21, 74, 52, 80, 98, 59, 12, 71, 85, 6, 34, 55, 82, 95, 63, 78, 15, 94, 60, 99, 76, 25, 40, 88, 0, 62, 96, 87, 51, 16, 18, 9, 19, 29, 45, 86, 53, 56, 31, 28, 61, 30, 33, 4, 67, 64, 58, 50, 54, 3, 13, 37, 27, 66, 77, 84, 69, 2, 41, 22, 92, 42, 44, 11, 36, 46, 79, 65, 72, 23, 17, 39, 5, 89, 35, 24, 83, 43, 57, 93, 32, 68, 26]
\end{quote}

\subsection{Integrated setting}
\label{app:integrated}
Here we train a ResNet-18 model for 100 epochs using the same optimizer as in \ref{app:stand-alone}, with the same batch size and data augmentations, with some exceptions. \textit{ER-ACE} starts with a learning rate of 0.01 and train on batch size of 10 examples as in the original paper, and all experiments on Split CUB-200 are conducted for 30 epochs with a batch size of 16 due to the dataset size and the resolution of its images. In experiments using the ArchCraft model, we used the ResAC-A model as implemented in the official paper's code.

\subsection{Baselines setting}
The setting for the experiments of the baseline methods is the same as in \ref{app:integrated} with some exceptions. For \textit{BiC}, the number of training epochs is 250, and the learning rate scheduler decays the learning rate by a factor of 0.1 on epochs 100, 150 and 200. For \textit{GEM}, the batch size is 32 and the learning rate starts from 0.03.For \textit{GDumb}, the batch size is 32. For \textit{iCaRL} the learning rate starts from 2, the weight decay is 0.00001 and the learning rate scheduler decays the learning rate by a factor of 0.2 on epochs 49 and 63.

\subsection{Compute resources}
\label{app:compute}
All experiments involved training deep learning models, necessitating the use of GPUs. For the Split CIFAR-100 experiments, 10 GB of GPU memory was used. The Split tinyImageNet experiments required 22 GB of GPU memory, while the Split CUB-200 experiments utilized 45 GB of GPU memory. Any other experiments conducted for the full research and not reported in the paper required the same compute resources.

\section{Additional Results}
\label{app:more_results}
\label{app:buffer_comp}

\begin{figure}[htb] 
\center

\begin{subfigure}{.25\textwidth}
\includegraphics[width=\linewidth]{figures/cifar100/legened_split_diff_herding.png}
\end{subfigure}
\\
\begin{subfigure}{.23\textwidth}
\includegraphics[width=\linewidth]{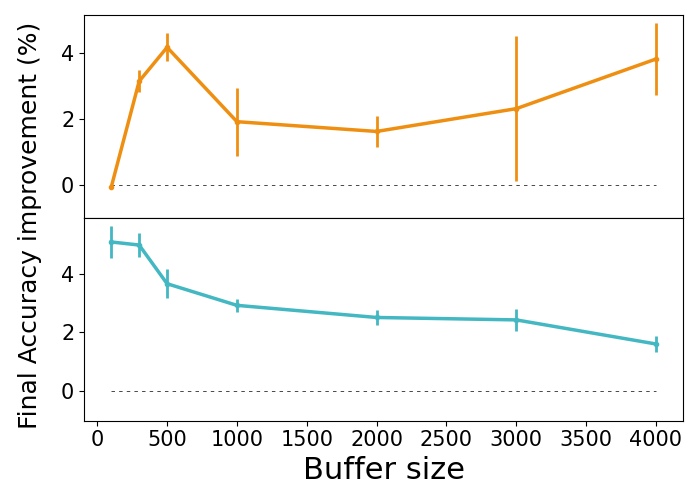}
\vspace{-0.3cm}
\caption{Split CIFAR-100}
\vspace{-0.1cm}
\end{subfigure}
\begin{subfigure}{.23\textwidth}
\includegraphics[width=\linewidth]{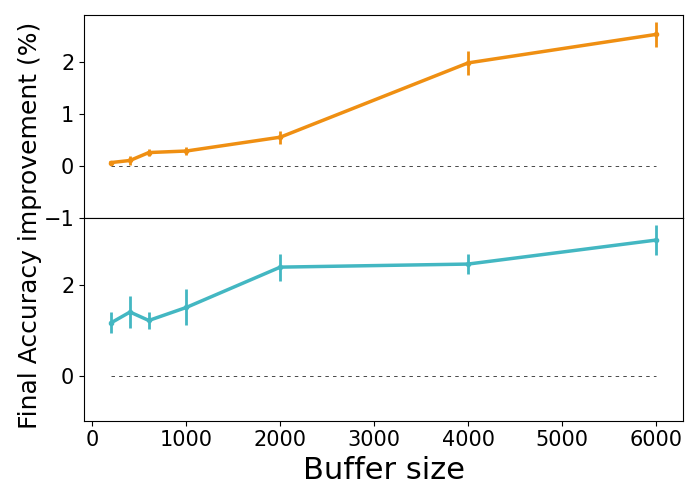}
\vspace{-0.3cm}
\caption{Split tinyImageNet}
\vspace{-0.1cm}
\end{subfigure}
\caption{Performance improvement of \MethodName\ when integrated with \textit{ER} and with \textit{ER-ACE} over various buffer sizes.}
\end{figure}

\begin{figure*}[thbp] 
\center
\begin{subfigure}{.7\textwidth}
\includegraphics[width=\linewidth]{figures/cifar100/legends_final_comp.png}
\end{subfigure}

\begin{subfigure}{.225\textwidth}
\includegraphics[width=\linewidth]{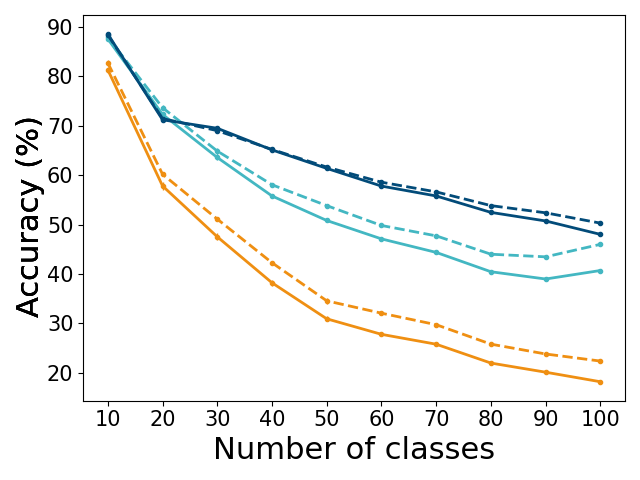}
\end{subfigure}
\hspace{.5cm}
\begin{subfigure}{.225\textwidth}
\includegraphics[width=\linewidth]{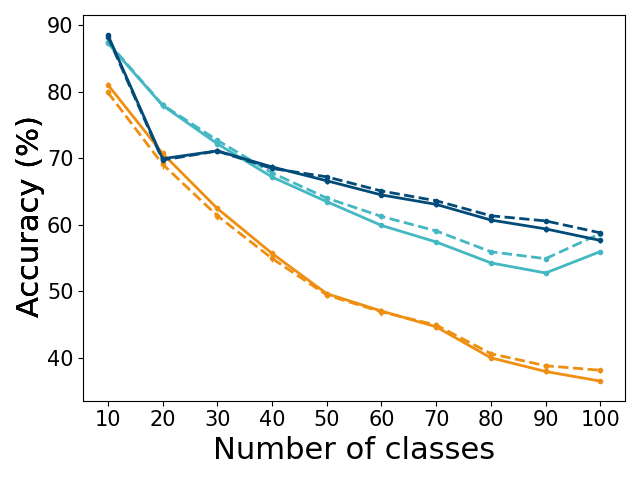}
\end{subfigure}
\hspace{.5cm}
\begin{subfigure}{.225\textwidth}
\includegraphics[width=\linewidth]{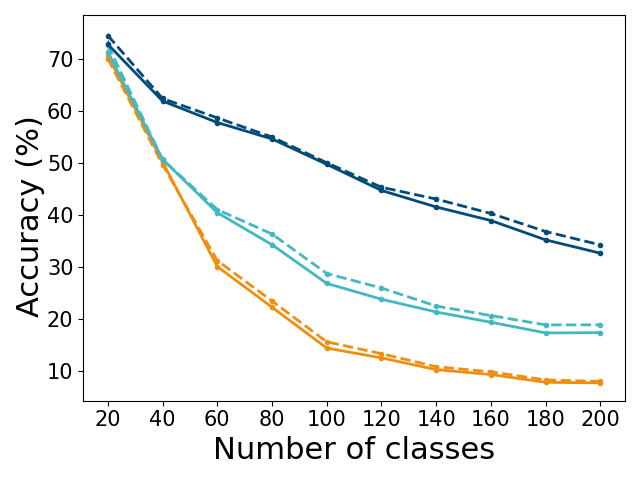}
\end{subfigure}
\\
\begin{subfigure}{.225\textwidth}
\includegraphics[width=\linewidth]{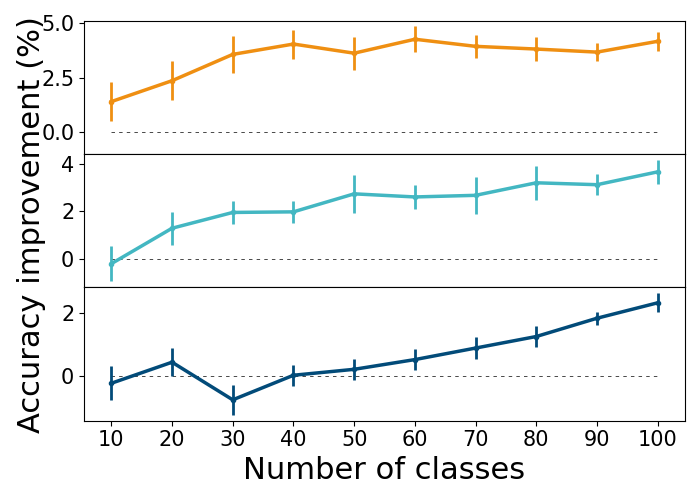}
\caption{Split CIFAR-100, $|\mem| = 500$} 
\end{subfigure}
\hspace{.5cm}
\begin{subfigure}{.225\textwidth}
\includegraphics[width=\linewidth]{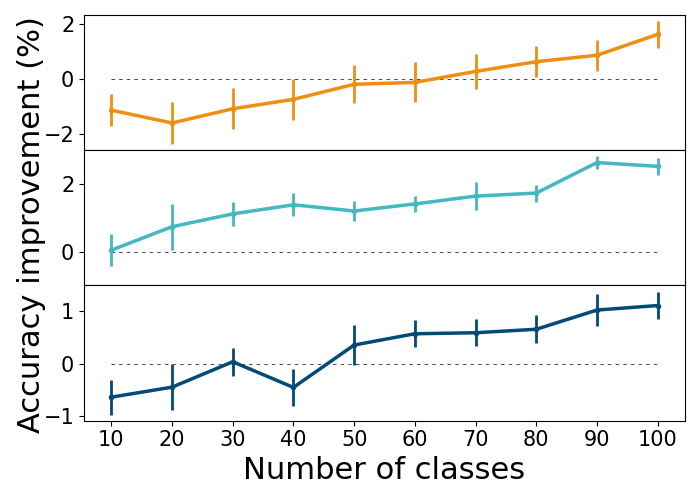}
\caption{Split CIFAR-100, $|\mem| = 2000$} 
\end{subfigure}
\hspace{.5cm}
\begin{subfigure}{.225\textwidth}
\includegraphics[width=\linewidth]{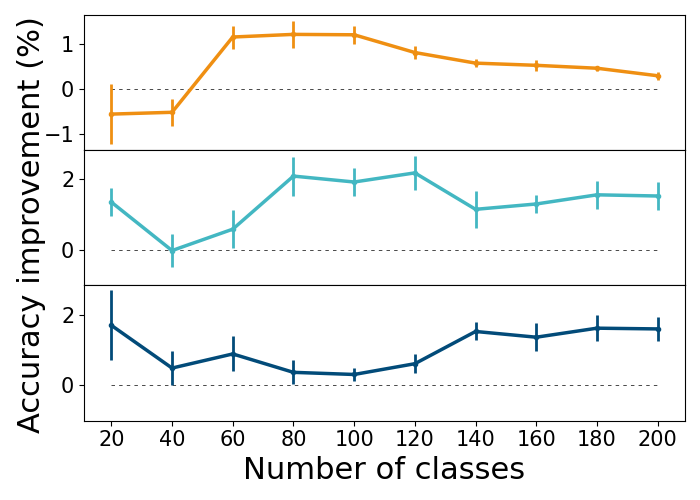}
\caption{Split tinyImageNet, $|\mem| = 1000$}
\end{subfigure}
\\
\begin{subfigure}{.225\textwidth}
\includegraphics[width=\linewidth]{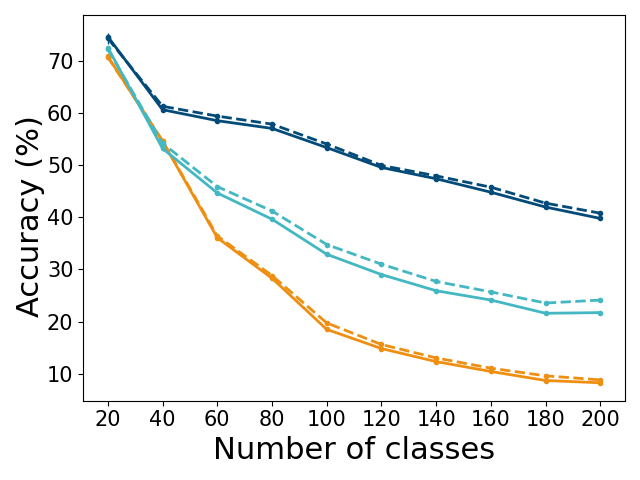}
\end{subfigure}
\hspace{.5cm}
\begin{subfigure}{.225\textwidth}
\includegraphics[width=\linewidth]{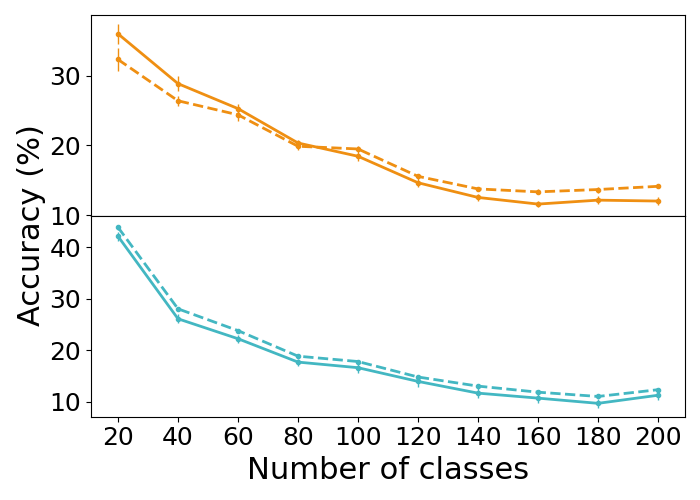}
\end{subfigure}
\\
\begin{subfigure}{.225\textwidth}
\includegraphics[width=\linewidth]{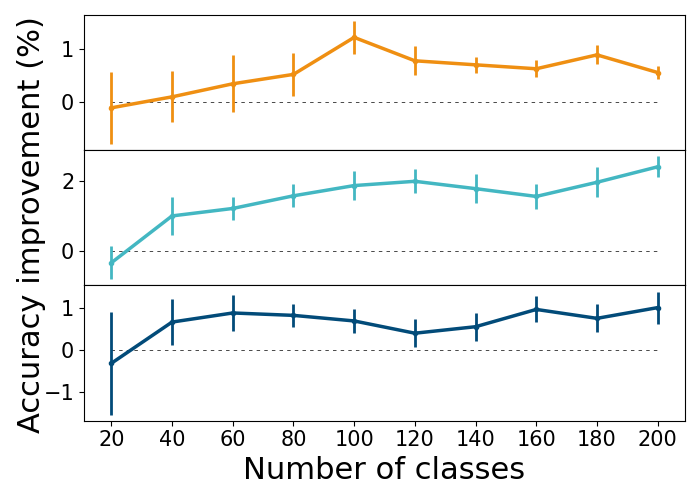}
\caption{Split tinyImageNet, $|\mem| = 2000$}
\end{subfigure}
\hspace{.5cm}
\begin{subfigure}{.225\textwidth}
\includegraphics[width=\linewidth]{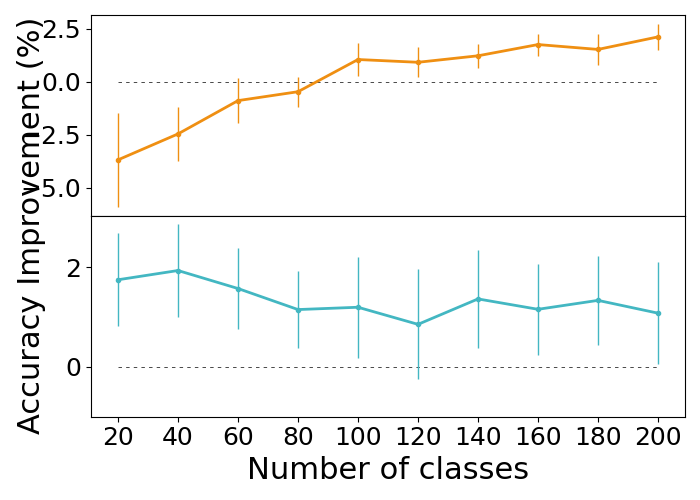}
\caption{Split CUB-200, $|\mem| = 400$} 
\end{subfigure}

\caption{The performance of ER-IL methods with and without \MethodName. First row displays the average accuracy after training incrementally on a different number of classes. Each color corresponds to a different ER-IL method, where the continuous line represents the vanilla method, while the dashed line represents the method with \MethodName\ as its selection strategy. The error bars correspond to standard error based on 4-10 repetitions. Second row depicts the difference in accuracy between \MethodName\ and another method (\textit{XDER}, \textit{ER-ACE}, and \textit{ER}) across all tasks.
}
\label{fig:accuracy_comp_app}
\vspace{-0.3cm}
\end{figure*}

\begin{figure*}[h] 
\center

\begin{subfigure}{.5\textwidth}
\includegraphics[width=\linewidth]{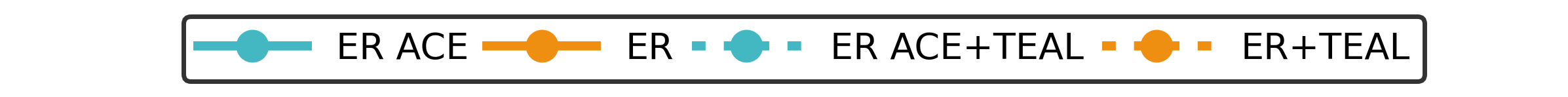}
\end{subfigure}
\\
\begin{subfigure}{.25\textwidth}
\includegraphics[width=\linewidth]{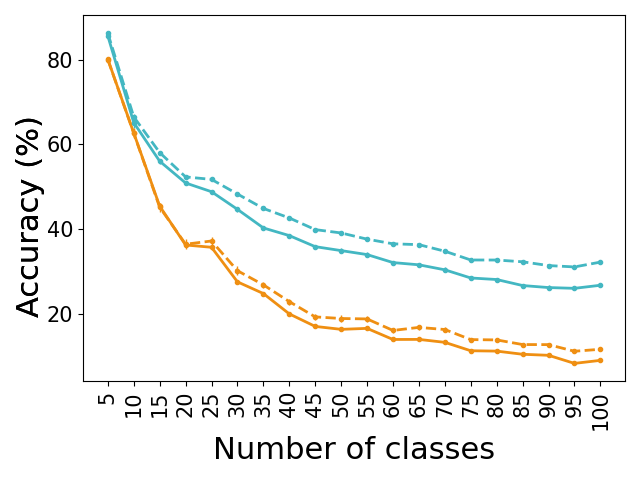}
\caption{$|\mem| = 300$} 
\end{subfigure}
\begin{subfigure}{.25\textwidth}
\includegraphics[width=\linewidth]{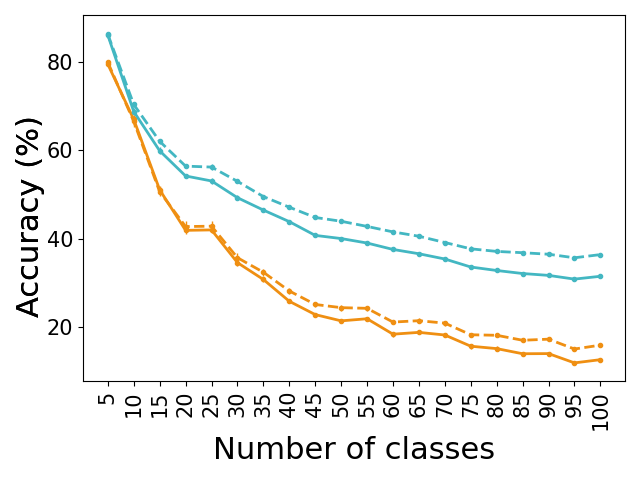}
\caption{$|\mem| = 500$} 
\end{subfigure}
\begin{subfigure}{.25\textwidth}
\includegraphics[width=\linewidth]{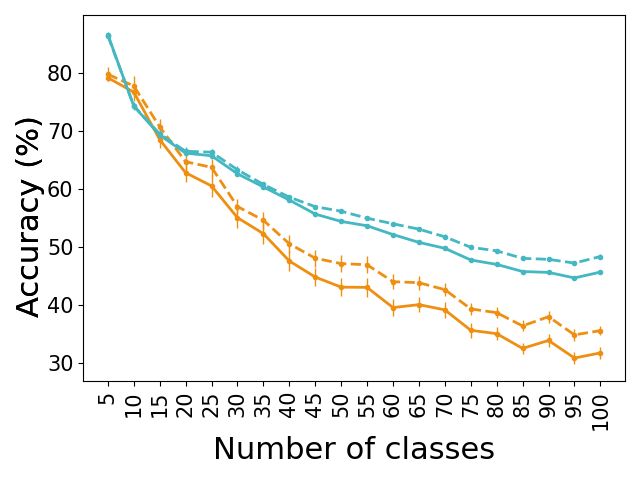}
\caption{$|\mem| = 2000$} 
\end{subfigure}

\caption{Split CIFAR-100: The performance of ER-IL methods with and without \MethodName\ with splitting CIFAR-100 into 20 tasks instead of 10. Each color corresponds to a different ER-IL method, where the continuous line represents the vanilla method, while the dashed line represents the method with \MethodName\ as its selection strategy. The error bars correspond to standard error based on 6 repetitions.}
\vspace{-0.3cm}
\end{figure*}

\label{app:vanila_comp}

\begin{figure}[H] 
\center
\begin{subfigure}{0.325\textwidth}
\includegraphics[width=\linewidth]{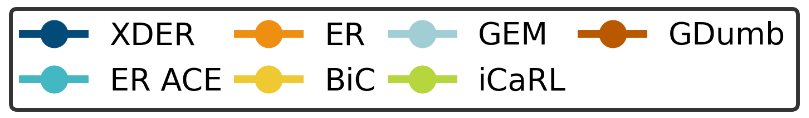}
\end{subfigure}
\begin{subfigure}{.23\textwidth}
\includegraphics[width=\linewidth]{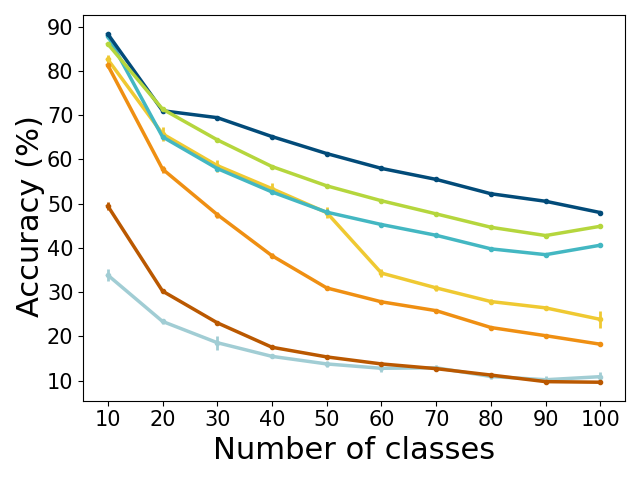}
\caption{$|\mem| = 500$} 
\end{subfigure}
\begin{subfigure}{.23\textwidth}
\includegraphics[width=\linewidth]{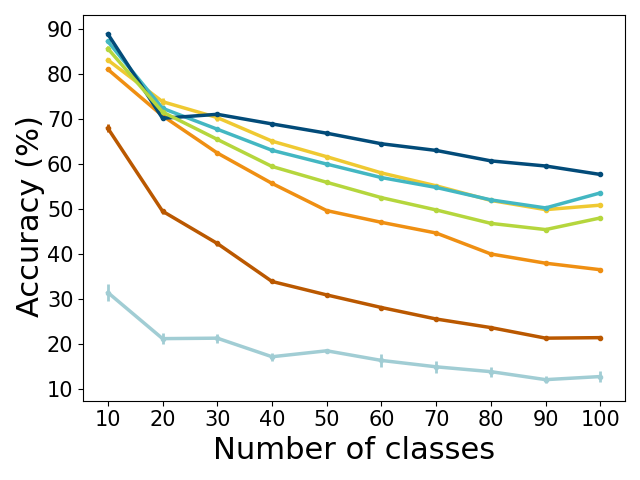}
\caption{$|\mem| = 2000$}
\end{subfigure}
\caption{Baseline CIL methods comparison on Split CIFAR-100 with different fixed buffer sizes. For each method, each point represents the average accuracy achieved on the corresponding task $t$, $A_{t}$.}
\label{fig:vanilla_comp_app}
\end{figure}

\end{document}